\documentclass{article}

\PassOptionsToPackage{numbers, compress}{natbib}

    \usepackage[preprint]{neurips_2025}

\usepackage[utf8]{inputenc} 
\usepackage[T1]{fontenc}    
\usepackage{hyperref}       
\usepackage{url}            
\usepackage{booktabs}       
\usepackage{amsfonts}       
\usepackage{nicefrac}       
\usepackage{microtype}      
\usepackage{xcolor}         

\usepackage{graphicx}
\usepackage{amsmath}
\usepackage{cleveref}
\usepackage{multirow}
\usepackage{subcaption}

\usepackage{graphicx}
\usepackage[normalem]{ulem}
\usepackage{multirow}
\useunder{\uline}{\ul}{}
\usepackage{placeins}   

\title{Mol-LLM: Multimodal Generalist Molecular LLM with Improved Graph Utilization}

\author{%
\textbf{Chanhui Lee\textsuperscript{\rm 1},}
\textbf{Hanbum Ko\textsuperscript{\rm 1},}
\textbf{Yuheon Song\textsuperscript{\rm 2},}
\textbf{Yongjun Jeong\textsuperscript{\rm 1}}\\
\textbf{Rodrigo Hormazabal\textsuperscript{\rm 3,4},}
\textbf{Sehui Han\textsuperscript{\rm 4},}
\textbf{Kyunghoon Bae\textsuperscript{\rm 4},}
\textbf{Sungbin Lim\textsuperscript{\rm 4,5 $\ast$}},
\textbf{Sungwoong Kim\textsuperscript{\rm 1}}\thanks{Corresponding Authors. \texttt{\{sungbin, swkim01\}@korea.ac.kr}.}
\\
\textsuperscript{\rm 1}Department of Artificial Intelligence, Korea University \\
\textsuperscript{\rm 2}Department of Artificial Intelligence, UNIST \\
\textsuperscript{\rm 3}Kim Jaechul Graduate School of AI, KAIST \\
\textsuperscript{\rm 4}LG AI Research \\
\textsuperscript{\rm 5}Department of Statistics, Korea University \\
}

\begin{document}
\maketitle
\begin{abstract}
Recent advances in large language models (LLMs) have led to models that tackle diverse molecular tasks, such as chemical reaction prediction and molecular property prediction.
Large-scale molecular instruction-tuning datasets have enabled sequence-only (e.g., SMILES or SELFIES) generalist molecular LLMs, and researchers are now exploring multimodal approaches that incorporate molecular structural information for further gains.
However, a genuinely multimodal, generalist LLM that covers a broad spectrum of molecular tasks has yet to be fully investigated.
We observe that naive next token prediction training ignores graph-structural information, limiting an LLM’s ability to exploit molecular graphs.
To address this, we propose (i) Molecular structure Preference Optimization (MolPO), which facilitates graph usage by optimizing preferences between pairs of correct and perturbed molecular structures, and (ii) an advanced graph encoder with a tailored pre-training strategy to improve the effect of graph utilization by MolPO.
Building on these contributions, we introduce Mol-LLM, the first multimodal generalist model that (a) handles a broad spectrum of molecular tasks among molecular LLMs, (b) explicitly leverages molecular-structure information, and (c) takes advantage of extensive instruction tuning.
Mol-LLM attains state-of-the-art or comparable results across the most comprehensive molecular-LLM benchmark—even on out-of-distribution datasets for reaction and property prediction, where it surpasses prior generalist molecular LLMs by a large margin.\footnote{The model, code, and data will be publicly available.}
\end{abstract}

\section{Introduction}
\label{sec:intro}
Large language models (LLMs)~\cite{GPT3, GPT4, Gemini, llama2} have been widely used to tackle diverse tasks across multiple domains, such as mathematics and code generation, by leveraging their broad knowledge base.
This achievement has recently motivated interest in applying LLMs to diverse molecular tasks—including molecular property prediction, chemical reaction prediction, description-guided molecule generation, and molecule captioning—all of which are essential in drug discovery and materials science~\cite{Yu2024LlaSMolAL, Pei2024BioT5TG, Fang2023MolInstructionsAL, Liu2023MolCAMG, Cao2023InstructMolMI, liu2024gitmol, zhang2024unimotunifiedmoleculetextlanguage, li20243dmoleculetextinterpretationlanguage}.
In particular, most molecular LLMs tend to leverage only one of the two key components for improved molecular language modeling, either molecular structure information or multitask instruction-tuning, rather than combining both.
Several studies~\cite{Liu2023MolCAMG, liu2024gitmol, zhang2024unimotunifiedmoleculetextlanguage} have moved away from conventional molecular language modeling based on 1D sequence such as SMILES~\cite{weininger1988} or SELFIES~\cite{Krenn_2020}, and have instead developed multimodal LLMs that incorporate 2D molecular graphs as an additional input modality, thereby representing molecular structures and topologies more faithfully while achieving better performance across diverse molecular tasks~\cite{liu2024multimodalmoleculestructuretextmodel,Yuyang_2022,su2022molecularmultimodalfoundationmodel}.
Meanwhile, other studies~\cite{Yu2024LlaSMolAL,Pei2024BioT5TG,Fang2023MolInstructionsAL,Cao2023InstructMolMI} have constructed instruction-tuning datasets for multiple molecular tasks and fine-tuned LLMs on these datasets. 
This approach enables the models to acquire transferable and generalizable knowledge, allowing them to understand and perform various tasks based on natural language instructions.

\begin{figure}[t!]\centering
      \begin{subfigure}[b]{0.38\columnwidth}
        \includegraphics[width=\linewidth]{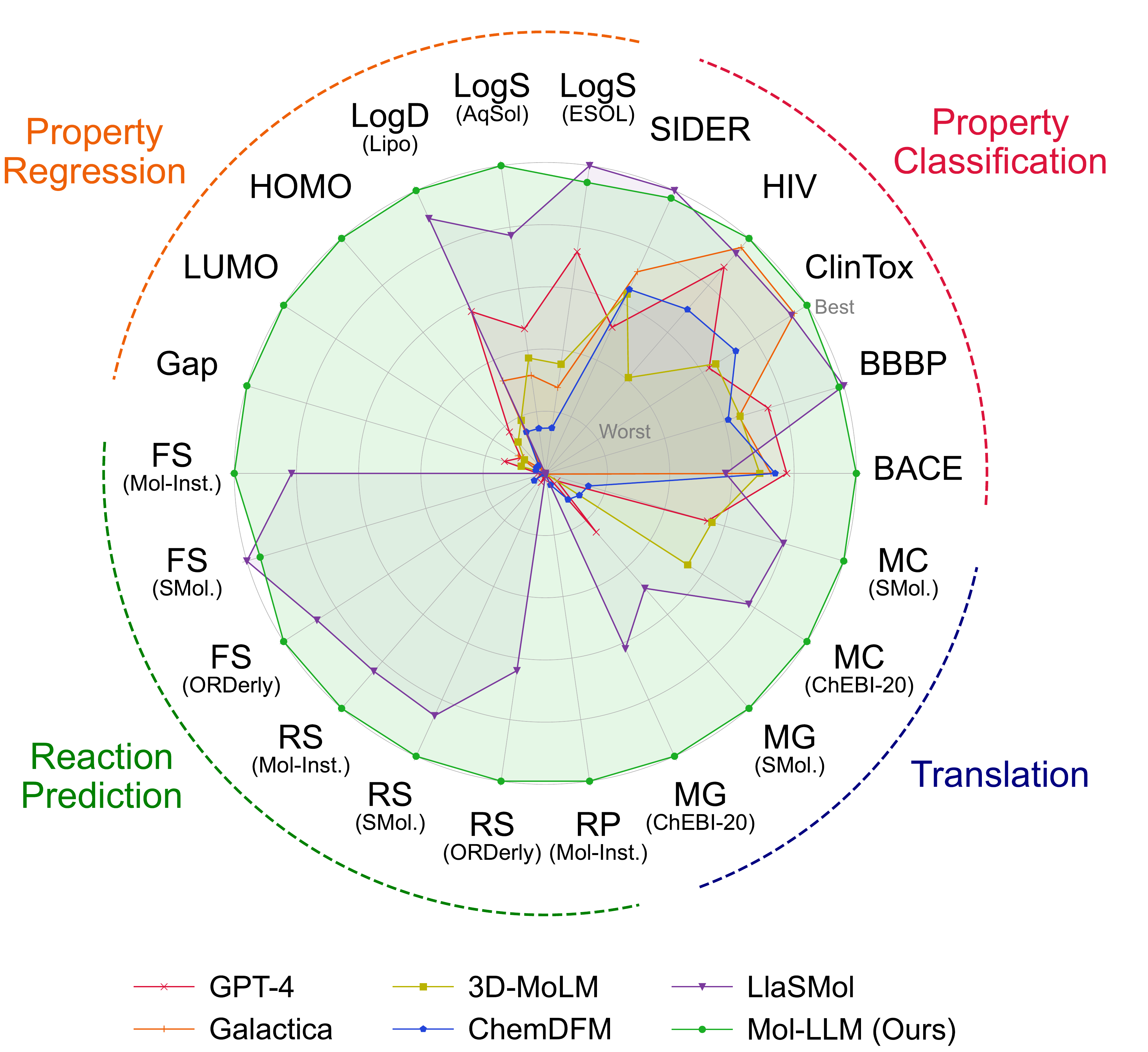}
        \label{fig:generalist}
      \end{subfigure}\hfill
      \begin{subfigure}[b]{0.52\columnwidth}
        \includegraphics[width=\linewidth]{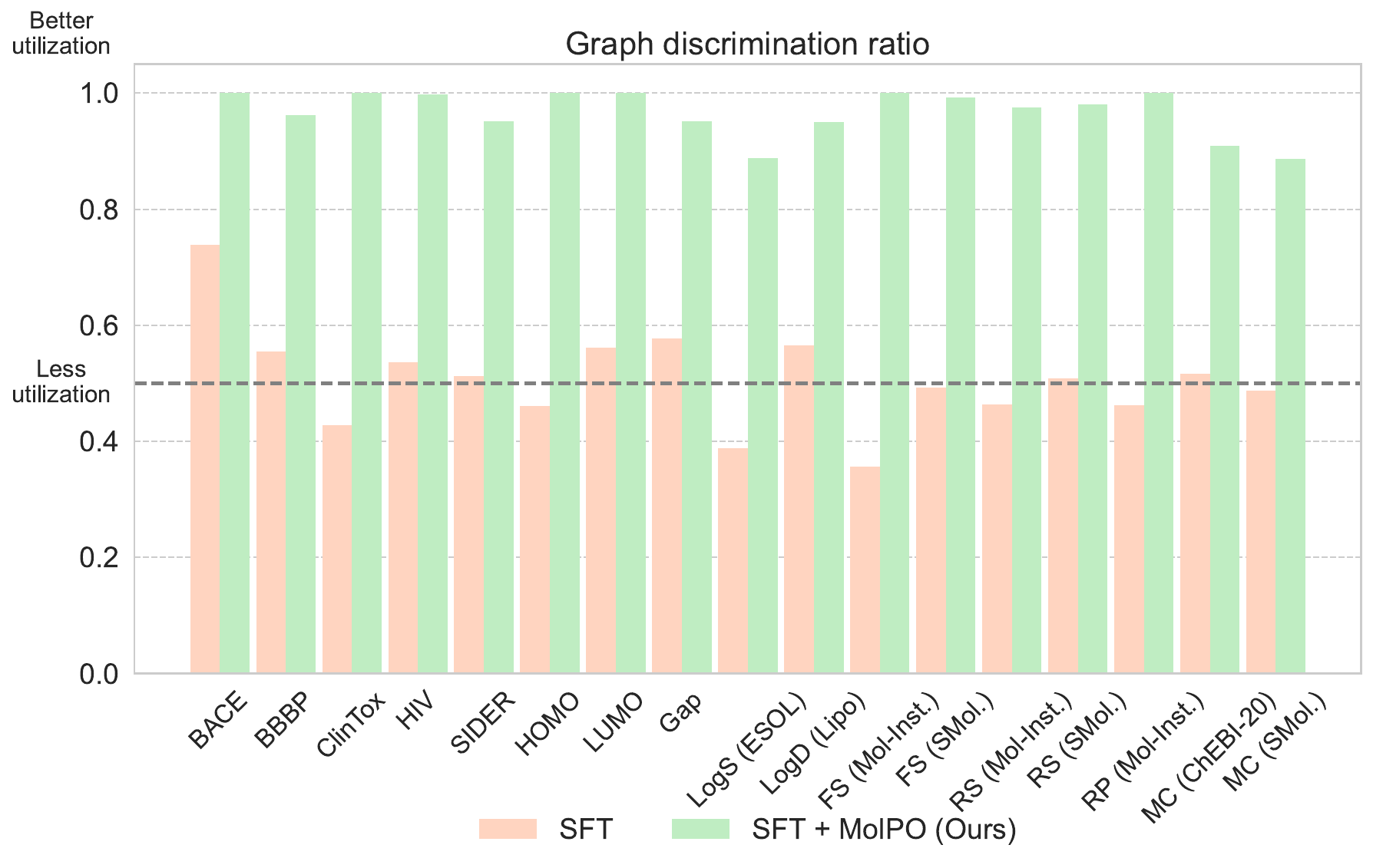}
      \end{subfigure}\hfill

    \caption{(Left) Performance comparison among generalist molecular LLMs with normalized primary metrics. (Right) Graph utilization comparison between SFT and proposed multimodal training (MolPO). Score closer to 1 indicate better use of graph, approaching 0.5 indicate less utilization.}
    \label{fig:graph-util}
\label{fig:generalist_graph_util}
\end{figure}

However, it is uncertain whether the multimodal molecular LLMs effectively use molecular structural information when trained with naive supervised fine-tuning (SFT).
To investigate this, we compare the likelihoods of the original and perturbed molecules, comparing how well the SFT model is at proper graph discrimination.
\Cref{fig:generalist_graph_util} shows that the SFT model hardly distinguishes between them on most molecular tasks, indicating that its molecular graph utilization is generally limited.
Moreover, despite the potential for synergistic performance improvements by molecular graph structure utilization and multitask instruction-tuning, few studies have fully harnessed the benefits of both approaches, especially for a universal molecular LLM.
Specifically, some recent studies~\cite{Cao2023InstructMolMI,zhang2024unimotunifiedmoleculetextlanguage,li20243dmoleculetextinterpretationlanguage,liang2023drugchatenablingchatgptlikecapabilities,pei20243dmolt5unified3dmoleculetext} have attempted to combine molecule graph structure information with instruction-tuning, however, their instruction-tuning focuses solely on task-specific fine-tuning.

In this paper, we propose a generalist molecular LLM, called Mol-LLM, that leverages multimodal molecule and extensive instruction-tuning, addressing the broadest range of molecular tasks.
In particular, while maintaining multimodal LLM architecture based on Q-Former \cite{li2023blip2bootstrappinglanguageimagepretraining}, we introduce a novel multimodal instruction-tuning based on Molecular structure Preference Optimization (MolPO), where the molecular LLM learns to optimize the molecular structural preferences between the pairs of the correct (chosen) molecular graph and the perturbed (rejected) molecular graph.
By creating rejected molecular graphs based on the substructures for molecular feature perturbation, the proposed MolPO mitigates the tendency to overlook graph information on various molecular tasks.
Additionally, to further increase the effect of molecular graph utilization by advanced representation on a wide variety of molecular distributions, we introduce a new graph neural network (GNN) pre-training strategy and architecture.
The proposed GNN pre-training framework combines two objectives: (i) functional group prediction, which teaches the model to accurately distinguish functional groups—the features that largely determine molecular properties—and (ii) SELFIES reconstruction, which helps the model preserve the molecular structure details from the molecular graph.
Upon GINE~\cite{hu2020strategies}, adopted by prior multimodal molecular LLMs~\cite{Liu2023MolCAMG, Cao2023InstructMolMI}, we incorporate a transformer-based GNN named TokenGT~\cite{kim2022pure}, to enhance the expressive power.
The resulting Mol-LLM shows strong performance and demonstrably better graph utilization on our benchmarks across a broad range of molecular tasks.
To the best of our knowledge, Mol-LLM is not only the first versatile generalist multimodal molecular LLM on a wide range tasks with a single generalist model, but it also surpasses other generalist models: LlaSMol~\cite{Yu2024LlaSMolAL}, ChemDFM~\cite{zhao2024chemdfm}, 3D-MoLM~\cite{li20243dmoleculetextinterpretationlanguage} on most bencharks as shown in~\Cref{fig:generalist_graph_util}, highlighting the power of graph modality synergized with extensive instruction-tuning.


In summary, our contributions are:
\begin{enumerate}
    \item \textbf{Mol-LLM}. We present Mol-LLM, which sets a new state-of-the-art on both in-distribution and out-of-distribution molecular benchmarks relative to existing generalist models.
    \item \textbf{Enhanced graph utilization}. To exploit 2D molecular graphs more effectively, we propose MolPO---a fine-tuning strategy that leverages perturbed molecules---alongside a GNN pre-training method and a hybrid graph encoder augmented with transformer architecture.
    \item \textbf{Extensive instruction-tuning}. We construct a large, molecule-focused instruction-tuning dataset and employ multimodal training to build a generalist model with significantly enhanced molecular understanding.
\end{enumerate}

\section{Mol-LLM: Multimodal Generalist Molecular Large Language Model}
\label{sec:method}
\begin{figure}[t]
  \centering
  \begin{subfigure}[b]{0.60\columnwidth}
    \includegraphics[width=\linewidth]{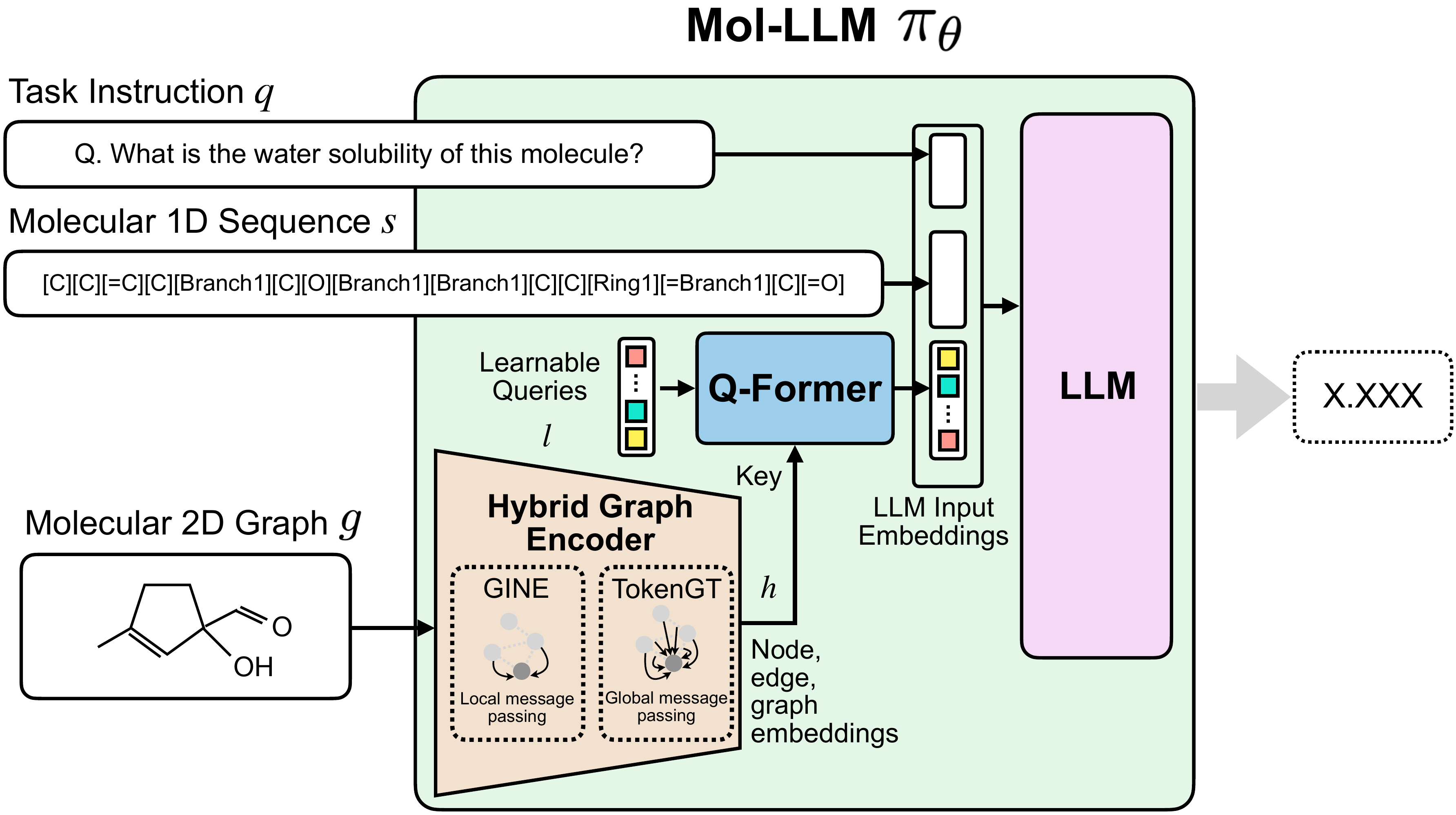}
  \end{subfigure}\hfill
  \begin{subfigure}[b]{0.39\columnwidth}
    \includegraphics[width=\linewidth]{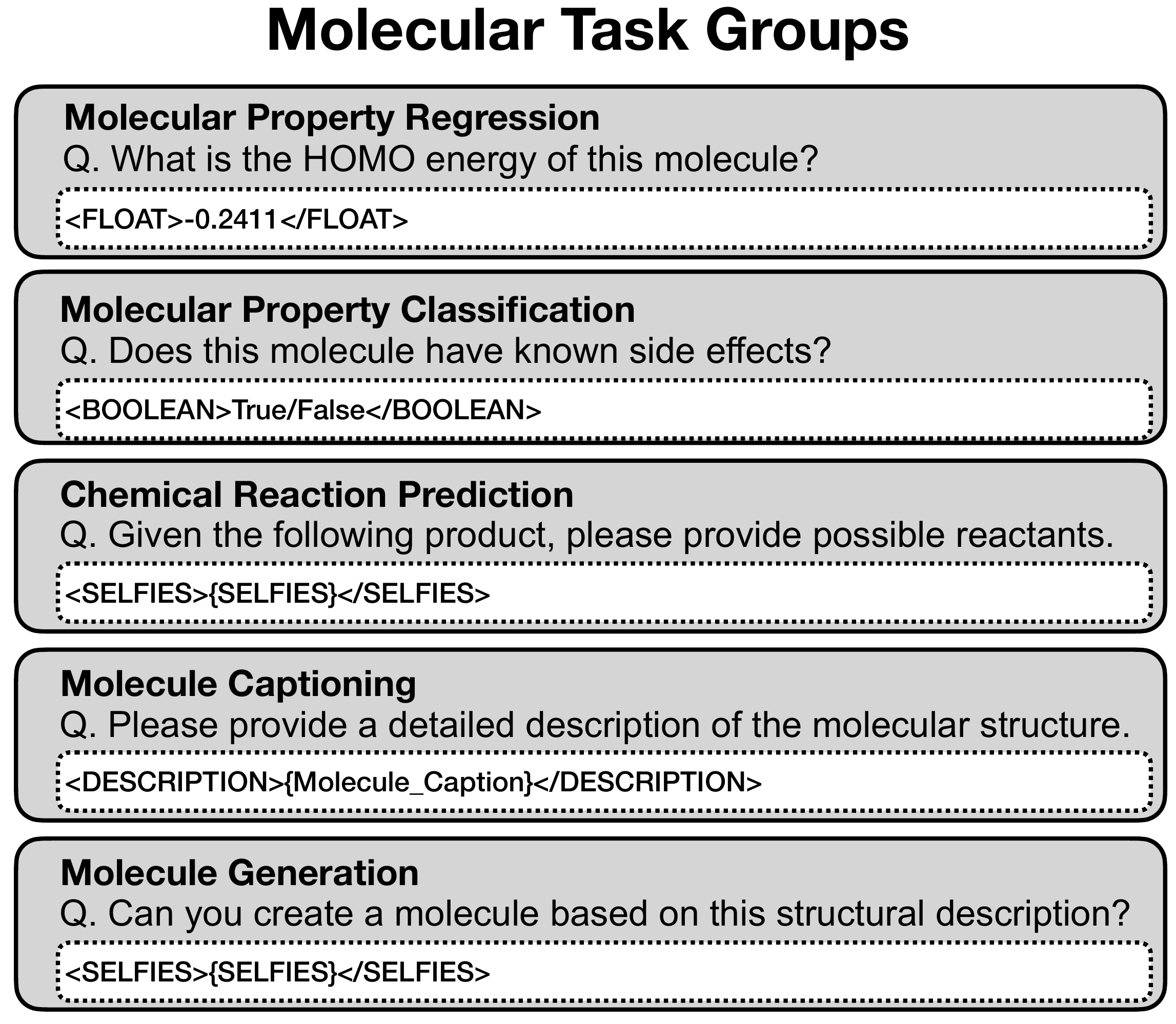}
  \end{subfigure}\hfill
  \caption{(Left) Overall structure of Mol-LLM. 
Molecular graph is encoded into a fixed-length token sequence by a hybrid graph encoder, followed by a Q-Former that outputs query embeddings to feed LLM, with corresponding task instruction and molecular 1D sequence. (Right) Representative downstream molecular tasks. 
}
  \label{fig:overview}
\end{figure}

\begin{figure}\centering
    \includegraphics[width=.98\textwidth]{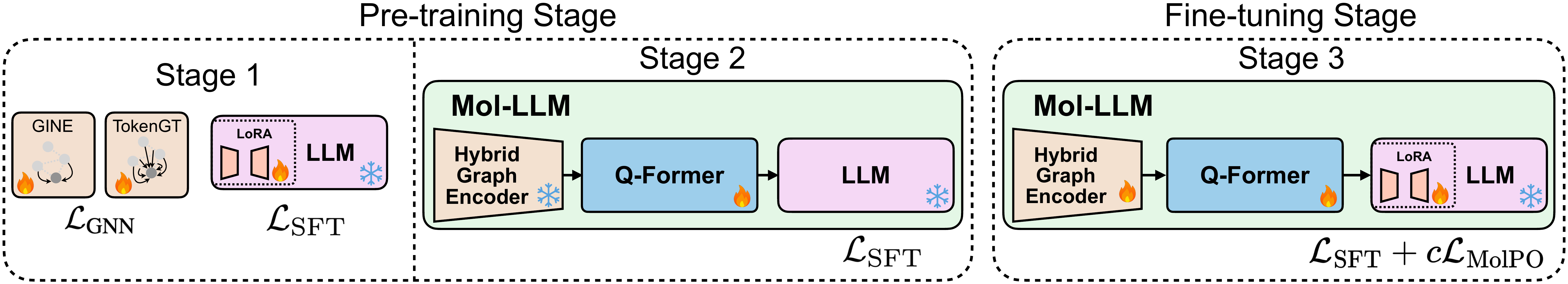}
    \caption{Overview of the three training stages and the loss function used at each stage. The training pipeline consists of a pre-training phase (Stages 1 and 2), followed by a fine-tuning phase (Stage 3). In Stage 1, all modules are trained independently and in parallel, whereas in Stages 2 and 3, the modules are trained in a unified architecture and loss function.}
    \label{fig:training_stages}
\end{figure}

This section introduces the model architecture, training strategy, and instruction-tuning dataset of Mol-LLM, a multimodal generalist molecular LLM.
As depicted in~\Cref{fig:overview}, Mol-LLM comprises a hybrid molecular graph encoder, a Q-Former for cross-modal projection between molecular graph and text, and a backbone LLM.
Utilizing the multimodal framework, the LLM addresses molecular task instructions and 1D molecular sequences directly, while feeding 2D molecular graph embeddings to the LLM through the hybrid graph encoder and Q-Former.
Such multimodal architectures are trained through three training stages, as depicted in \Cref{fig:training_stages}.

\subsection{Model Architecture}
\label{subsec:architecture}
\paragraph{Hybrid Graph Encoder}
\label{paragraph:graph_encoder}
Previous studies on multimodal molecular LLMs using 2D molecular graphs \cite{Liu2023MolCAMG, Cao2023InstructMolMI} have adopted the GINE architecture~\cite{hu2020strategies}, since it captures local graphical structure efficiently.
However, addressing diverse molecular tasks across various data distributions requires the ability to process large molecules as well. 
This consideration led us to the simultaneous usage of TokenGT~\cite{kim2022pure} as a graph encoder, which is designed to enhance global context understanding and mitigate over-smoothing~\cite{li2018deeper} in large graphs via a transformer architecture.
For a 2D molecular graph $G=(V,E)$, the GINE encoder $f^{G}$ outputs a graph embedding $h_g^{G}\in\mathbb{R}^{1 \times d_g}$ and node embeddings $h_v^{G}\in\mathbb{R}^{|V|\times d_g}$, where $d_g$ is the embedding dimension.
Otherwise, the TokenGT encoder $f^{T}$ outputs not only a graph embedding $h_g^{T}\in\mathbb{R}^{1 \times d_g}$ and node embeddings $h_v^{T}\in\mathbb{R}^{|V|\times d_g}$, but also edge embeddings $h_e^{T}\in\mathbb{R}^{|E|\times d_g}$.
We then concatenate all the embeddings obtained by both encoders $h_g^{G}, h_v^{G}, h_g^{T}, h_v^{T}$, and $h_e^{T}$ along the first dimension to obtain $h \in \mathbb{R}^{(2|V| + |E| + 2)\times d_g}$, which is then used as the key for the Q-Former.

\paragraph{Cross-modal Projector (Q-Former)}
\label{paragraph:cross_modal_projector}
Querying transformer (Q-Former)~\citep{li2023blip2bootstrappinglanguageimagepretraining} is a modality-bridging transformer that converts the varying number of concatenated embeddings for each molecular graph into a fixed-length token sequence, enabling efficient batch processing.
Specifically, structural information is distilled via cross-attention between $N_q = 32$ learnable query vectors $l \in \mathbb{R}^{32 \times d_q}$, initialized randomly, and the concatenated molecular embeddings $h \in \mathbb{R}^{(2|V| + |E| + 2) \times d_g}$, producing 32 tokens aligned with the text modality.
The 32 tokens are concatenated with the task instruction as well as the SELFIES string before being fed to the LLM.

\paragraph{Backbone Large Language Model}
\label{paragraph:LLM}
We adopt Mistral‑7B‑Instruct‑v0.3 \citep{jiang2023mistral7b} as a backbone LLM, following \citet{Yu2024LlaSMolAL}. 
In order to improve the efficiency in solving molecular tasks, we extend the token codebook with the 3K SELFIES vocabulary from BioT5+ \citep{Pei2024BioT5TG} and add dedicated tokens for the digits 0–9, the decimal point, and the negative sign, thereby enabling direct number prediction for regression tasks.
Additional task‑specific vocabulary covers boolean labels, textual descriptions, and reaction routes, allowing Mol-LLM to natively produce the heterogeneous answer formats required by downstream applications.
Examples of these extra tokens appear on the right side of~\Cref{fig:overview}.

\subsection{Multimodal Training}
\label{subsec:pre-training_stage}

\begin{figure}[t]
  \centering
  \begin{subfigure}[b]{0.62\columnwidth}
    \includegraphics[width=\linewidth]{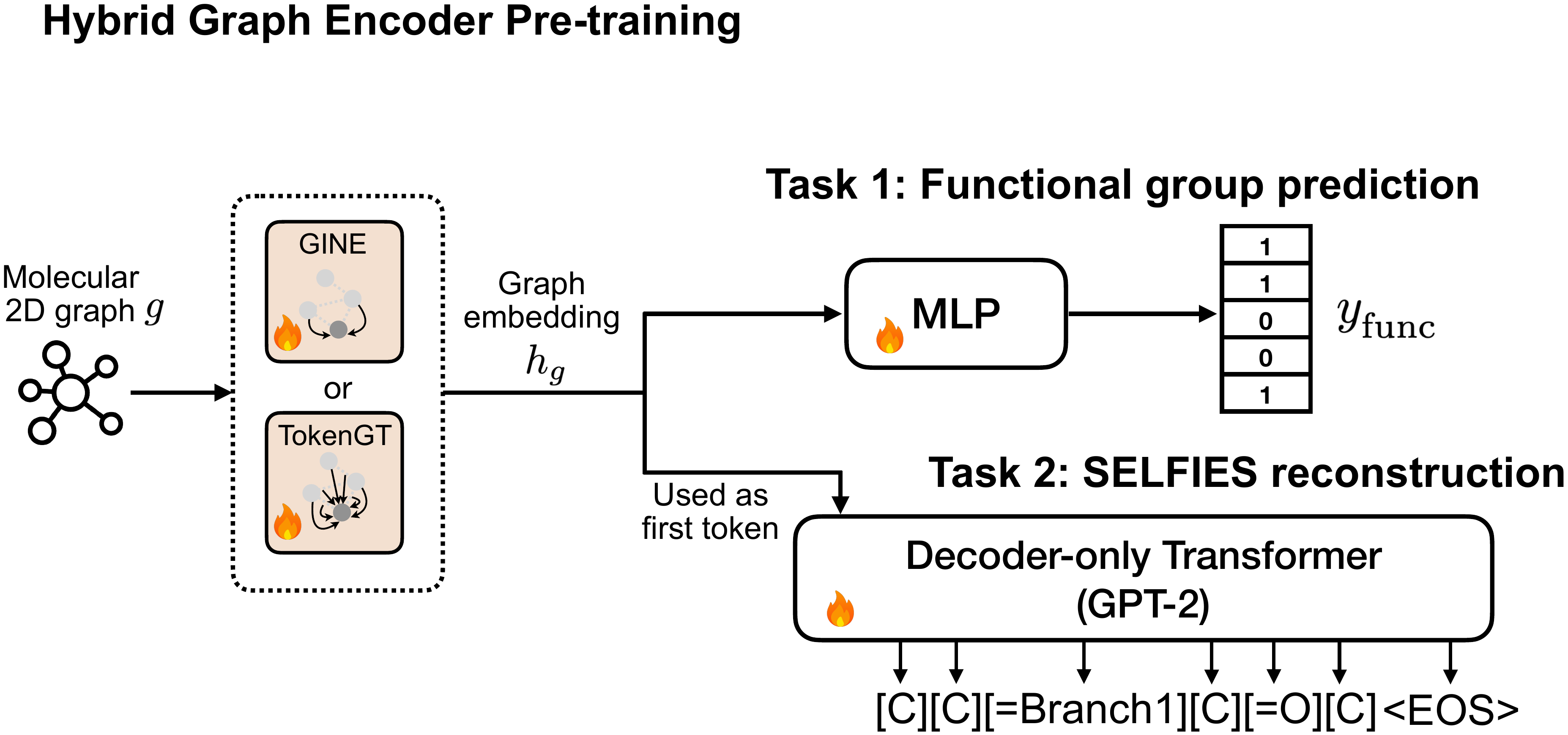}
  \end{subfigure}\hfill
  \begin{subfigure}[b]{0.37\columnwidth}
    \includegraphics[width=\linewidth]{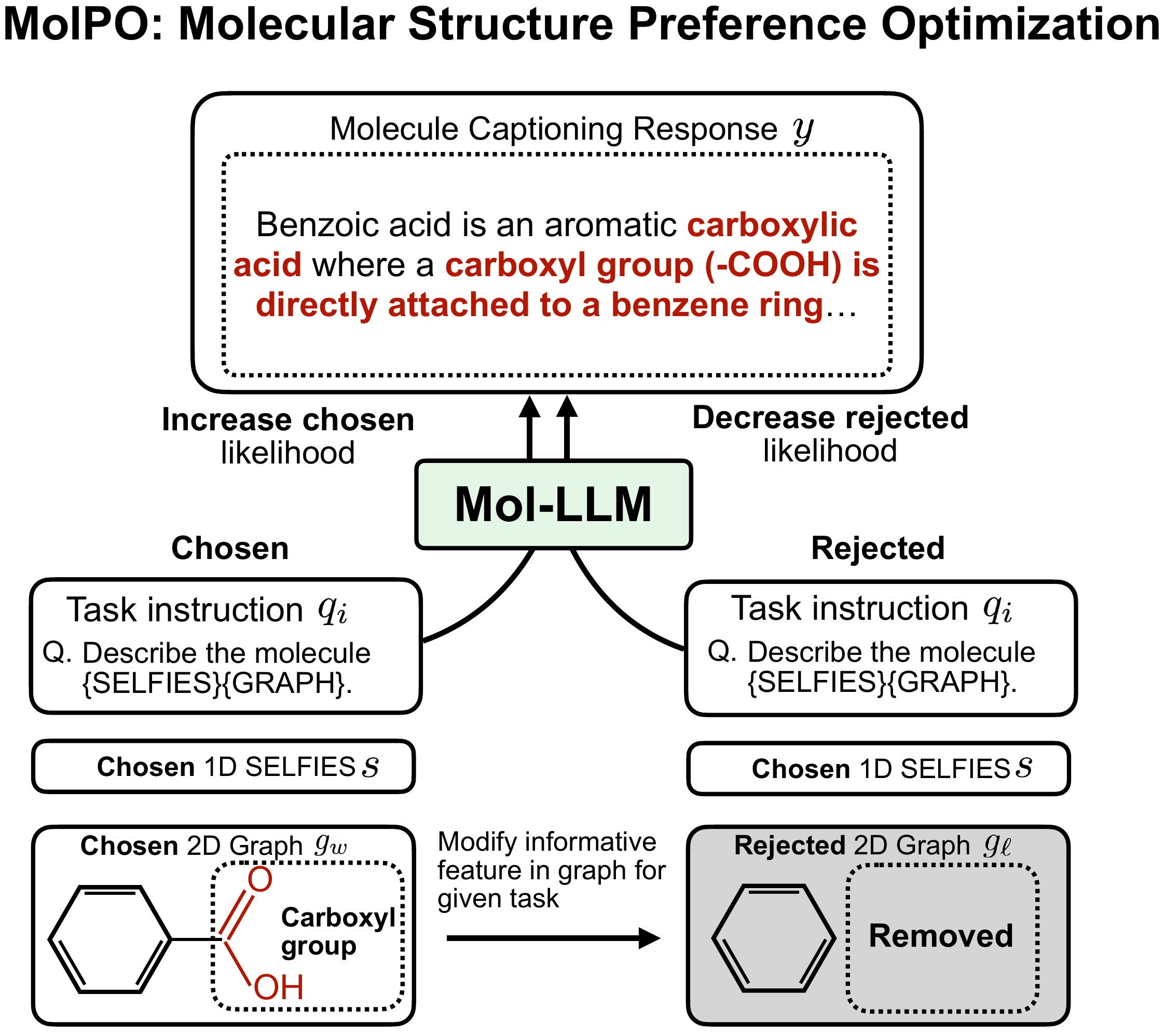}
  \end{subfigure}\hfill
  \caption{(Left) Overview of the two graph pre-training tasks for the proposed hybrid graph encoder. 
  Two distinct GNN backbones, GINE and TokenGT, are trained independently. 
  (Right) Illustration of the MolPO training objective, which contrasts a chosen molecule with a rejected molecule.}
  \label{fig:gnnpt_plus_molpo}
\end{figure}

\paragraph{Stage 1 - Graph Encoder and LLM Pre-training}
\label{paragraph:stage1}
The hybrid graph encoder comprises two GNNs, GINE and TokenGT.
We pre-train these two GNNs in parallel with the LLM.
The GNN pre-training comprises two complementary tasks: functional group prediction and SELFIES reconstruction.
Functional group prediction strengthens representations of the functional groups that govern molecular properties, whereas SELFIES reconstruction encourages the encoder to preserve global structural information.
Both tasks share the same graph-level embedding~$h_g$ produced by the GNN, as illustrated on the left side of~\Cref{fig:gnnpt_plus_molpo}.
After discarding extremely common or extremely rare functional groups, we retain $K=72$ distinct groups (dataset construction details are given in~\Cref{appx:functional_group_prediction_dataset_construction}).
For functional group prediction, $h_g$ is passed through a three-layer MLP ($1024\rightarrow1024\rightarrow72$) $f_\theta^\text{MLP}$ and trained with the binary cross-entropy loss
$
\mathcal{L}_{\mathrm{func}}
  = -\sum_{k=1}^{K} \Bigl(
      y_{\mathrm{func}}^{(k)} \log f_\theta^\text{MLP}(h_g)^{(k)}
      + \bigl(1 - y_{\mathrm{func}}^{(k)}\bigr)
        \log \bigl(1 - f_\theta^\text{MLP}(h_g)^{(k)}\bigr)
    \Bigr)
$
, where superscript $(k)$ is the value for functional group $k$.
SELFIES reconstruction reuses $h_g$ as a context for a GPT‑2 decoder $\pi_\theta^\text{GPT-2}$ that learns to reproduce molecule's SELFIES string $s$:
$
\mathcal{L}_{\mathrm{recon}}
  = -\sum_{t} \log \pi_\theta^\text{GPT-2}\bigl(s_t \mid h_g,\, s_{<t}\bigr).
$
The graph encoder is optimized with the combined loss
$
\mathcal{L}_{\mathrm{GNN}}
  = \mathcal{L}_{\mathrm{func}} + \mathcal{L}_{\mathrm{recon}}.
$
Additional training procedures and hyperparameters are provided in~\Cref{appx:details_of_graph_encoder_pre_training}.

The LLM pre-training serves two purposes: (i) injecting molecule-specific prior knowledge and (ii) reducing the compute required during later multimodal training.
Accordingly, we pre-train the LLM on exactly the same dataset that will be used later for fine-tuning, optimizing a token-level cross-entropy objective.
Given a training instance consisting of a task instruction~$q$, a molecular SELFIES string~$s$, and a ground truth answer~$y$, we minimize $\mathcal{L}_{\mathrm{SFT}} = -\sum_{t} \log \pi_\theta^\text{LLM}\bigl(y_t \mid s, q, y_{<t}\bigr)$ where $t$ indexes tokens.


\paragraph{Stage 2 - Q-Former Pre-training}
\label{paragraph:stage2}
In Stage 2, only the Q-Former is updated, while both the GNN and the LLM remain frozen.
Following~\citet{llava}, we simply reuse the fine‑tuning dataset, in which molecular representations and natural language tokens appear in an interleaved format.
For each training instance $(s, q, y)$, the SELFIES string~$s$ is converted into its corresponding molecular graph~$g$.
The combined model $\pi_\theta$ (GNN+Q-Former+LLM) is then trained for one epoch with the loss defined as
$
\mathcal{L}_{\mathrm{SFT}}
= -\sum_{t} \log \pi_\theta\bigl(y_t \mid s, q, g, y_{<t}\bigr)
$.

\paragraph{Stage 3 - MolPO: Molecular Structure Preference Optimization}
We observed that using only SFT training as in conventional multimodal Molecular LLMs \citep{Liu2023MolCAMG, Cao2023InstructMolMI, pei20243dmolt5unified3dmoleculetext, li20243dmoleculetextinterpretationlanguage}, result in a graph bypass phenomenon (\Cref{fig:graph-util}) in solving molecular tasks.
To resolve the graph bypass issue, we propose Molecular structure Preference Optimization (MolPO).
Rather than simply inputting multimodal molecules into the LLM without consideration for multimodal utilization, MolPO promotes the practical utilization of multimodal molecules by learning the preferences between an original (chosen) graph $g_w$ and a perturbed (rejected) graph $g_\ell$, which is inspired by mDPO~\cite{Wang2024mDPOCP}.
Constructing $g_\ell$, it is crucial to introduce perturbations that alter the relationship between the graph and the target.
Since molecular features can generally be identified on a substructure basis, we substitute the substructures found in the original $g_w$ as in \Cref{fig:gnnpt_plus_molpo} right panel.
This approach offers the advantage of being applicable to any molecular task without significant computational costs or requiring task-specific graph perturbation design (details in \Cref{appx:gen_preference}).
Based on the reward formulation $r_{w,i}=\frac{\beta}{|y|}\sum_t\log\pi_\theta(y_t\mid g_w,s,q_i,y_{<t})$ and $r_{\ell,i}=\frac{\beta}{|y|}\sum_t\log\pi_\theta(y_t\mid g_\ell,s,q_i,y_{<t})$ motivated from SimPO~\cite{Meng2024SimPOSP}, the MolPO objective is defined as follows:
\begin{gather}
    \mathcal{L}_{\text{MolPO}}=\mathbb{E}_{(s,q_i,g,y) \sim \mathcal{D}_\text{tr}}[-\log\sigma\bigl(\min(r_{w,i}-r_{\ell,i},\lambda_{\text{clip}}|r_{w,i}|)-\gamma_i\bigr)],
\end{gather}
where $\lambda_\text{clip}$ is coefficient, and $\mathcal{D}_\text{tr}$ denotes training dataset.  $\gamma_i=\lambda_\text{margin}|\mathbb{E}_{(g_w, s, q_i, y)}[r_{w, i}]|$ is a task-adaptive target reward margin for each $i$-th molecular task, calculated during training.
The entire training objective combined is $\mathcal{L}_\text{SFT} + c\mathcal{L}_\text{MolPO}$, where $c$ is a constant.

In developing a generalist over diverse molecular tasks, we experimentally observed that adopting SimPO's task-agnostic target reward margin $\gamma$ results in inappropriate log-sigmoid values due to highly variant reward orders of magnitude across tasks. 
However, modeling the task-specific reward scale as a hyperparameter is not an ideal solution either, as it adds an additional burden of hyperparameter search for each molecular task. 
Instead, we introduce a task-adaptive target reward margin with only a task-agnostic hyperparameter $\lambda_\text{margin}$, where the expectation is estimated using an exponential moving average during training.

In addition, given that it is generally easier to lower the rejected reward than to increase the true chosen reward, the preference reward margin can be empirically manipulated by simply reducing $r_{\ell,i}$ without a corresponding enhancement of $r_{w,i}$.
To fully harness the benefits of preference optimization without the drawbacks, we introduce margin clipping to appropriately control the influence of the reward margin on parameter updates. 
Specifically, the margin is constrained so that it cannot exceed a fraction, $\lambda_{\text{clip}}$ of $|r_{w,i}|$. 
Through this simple margin clipping, the model is prevented from the circumventing unintended effect of preference optimization by solely reducing the rejected reward. Further details of MolPO training are provided in~\Cref{appx:details_of_molpo_objective}.

\subsection{Extensive Instruction-tuning Dataset}
\label{sec:extensive_instruction_tuning}
The instruction-tuning dataset for Mol-LLM spans five major molecular task groups: property regression, property classification, reaction prediction, description-guided molecule generation, and molecule captioning.
Property regression consists of five tasks---LogS for water solubility (ESOL~\cite{Yu2024LlaSMolAL}), LogD for lipophilicity (Lipo~\cite{Yu2024LlaSMolAL}), HOMO~\cite{Fang2023MolInstructionsAL}, LUMO~\cite{Fang2023MolInstructionsAL}, and HOMO-LUMO gap~\cite{Fang2023MolInstructionsAL}, and property classification comprises BACE~\cite{Pei2024BioT5TG}, BBBP~\cite{Yu2024LlaSMolAL}, ClinTox~\cite{Yu2024LlaSMolAL}, HIV~\cite{Yu2024LlaSMolAL}, SIDER~\cite{Yu2024LlaSMolAL}.
Reaction prediction covers forward synthesis (FS), retrosynthesis (RS), and reagent prediction (RP), with FS and RS each divided into Mol-Instructions~\cite{Fang2023MolInstructionsAL} and SMolInstruct~\cite{Yu2024LlaSMolAL} subsets according to their dataset sources.
The description-guided molecule generation and molecule captioning tasks are similarly split into ChEBI-20~\cite{Edwards2022TranslationBM} and SMolInstruct based on their origins.
In addition, to enhance the understanding of IUPAC~\cite{iupac}—frequently used in molecular text captions—we incorporate an IUPAC and SELFIES translation dataset~\cite{Yu2024LlaSMolAL} to construct an 3.3M extensive instruction-tuning dataset (details in~\Cref{appx:dataset_ind}).
\section{Experiments}
\label{sec:experiments}
\subsection{Experimental Setup}
\label{subsec:experimental_setup}

\paragraph{Baseline Models}
We group the molecular LLMs compared with Mol-LLM into three broad categories.
Specialist models are trained for a single molecular task; semi-generalist models cover a specific task group within one model but do not span all task groups; and generalist models are designed to handle every molecular task group.
Representative examples are MolCA~\cite{Liu2023MolCAMG} for the specialist category, BioT5+~\cite{Pei2024BioT5TG} for the semi-generalist category, and Galactica~\cite{galactica} and LlaSMol~\cite{Yu2024LlaSMolAL} for the generalist category.
Comprehensive details on all baseline models can be found in~\Cref{appx:baseline_models}.

\paragraph{Evaluation Benchmark}
In addition to the molecular tasks described in \Cref{sec:extensive_instruction_tuning}, we evaluate molecular LLM robustness to out-of-distribution (OOD) by proposing two evaluation benchmarks.
For LogS prediction, we retain high-confidence solubility labels from AqSol~\cite{Sorkun2019AqSolDBAC}, exclude every molecule that also appears in ESOL, and collect molecules of high consistency among labels to construct the OOD evaluation versus ESOL.
For reaction prediction, we gather 23K FS and 59K RS data instances from the ORDerly~\cite{Wigh2024ORDerlyDS} repository except USPTO~\cite{Wei2010ANM_uspto}, apply a scaffold split to remove motif overlap with Mol-Instructions~\cite{Fang2023MolInstructionsAL} and SMolInstruct~\cite{Yu2024LlaSMolAL}, and reserve 5K examples for evaluation in each task.
Full OOD dataset construction details are provided in Appendix~\ref{appx:dataset_ood}.

\paragraph{Evaluation Metrics}
For property prediction tasks, we report the root mean squared error (RMSE) or mean absolute error (MAE) in regression, and in classification tasks, receiver operating characteristic area under the curve (ROC-AUC) using the predicted probability of the positive class (i.e., \emph{True} token).
For reaction prediction and description-guided molecule generation, we evaluate exact match with the target molecule (EXACT), textual similarity (BLEU)~\cite{papineni2002bleu}, molecular fingerprint similarity based on RDKit~\cite{landrum2013rdkit}, MACCS keys~\cite{durant2002reoptimization}, and Morgan~\cite{morgan1965generation} fingerprints (RDK FTS, MACCS FTS, and MORGAN FTS, respectively), and the proportion of generated molecules that are chemically valid (VALIDITY).
For the molecule captioning task, we measure similarity between the generated and reference descriptions using BLEU-2, BLEU-4, ROUGE-1~\cite{lin2004rouge}, ROUGE-2, ROUGE-L, and METEOR~\cite{banerjee2005meteor}.
For a comprehensive assessment of the molecular LLM, we analyze a wide range of metrics for each molecular task. 
However, due to space limitations, the main paper reports only the primary metrics.
The complete results are provided in~\Cref{appx:full_exp_results}).

\subsection{Results}
\label{subsec:results}
\begin{table}[t!]
\footnotesize\centering\setlength{\tabcolsep}{4.4pt}
\caption{
Performance comparison on molecular property prediction tasks from the MoleculeNet~\cite{Wu2017MoleculeNetAB} benchmark. 
A superscript * indicates results evaluated with an official checkpoint, and "NA" denotes cases where no official checkpoint is available.
\textbf{Boldface} highlights the best scores among generalist models.
For semi-generalist models, each variant is annotated with the task group on which it is trained. 
GPT-4 is evaluated with 5-shots, except for classification performances borrowed from \citet{zhao2024chemdfm} with zero-shot.
}
\resizebox{\textwidth}{!}{
\begin{tabular}{@{}lrrrrrrrrrr@{}}
\toprule
Task &
  \multicolumn{1}{c}{LogS} &
  \multicolumn{1}{c}{LogD} &
  \multicolumn{1}{c}{HOMO} &
  \multicolumn{1}{c}{LUMO} &
  \multicolumn{1}{c}{Gap} &
  \multicolumn{1}{c}{BACE} &
  \multicolumn{1}{c}{BBBP} &
  \multicolumn{1}{c}{ClinTox} &
  \multicolumn{1}{c}{HIV} &
  \multicolumn{1}{c}{SIDER} \\ \midrule
Metric &
  \multicolumn{1}{c}{RMSE (↓)} &
  \multicolumn{1}{c}{RMSE (↓)} &
  \multicolumn{1}{c}{MAE (↓)} &
  \multicolumn{1}{c}{MAE (↓)} &
  \multicolumn{1}{c}{MAE (↓)} &
  \multicolumn{1}{c}{ROC-AUC (↑)} &
  \multicolumn{1}{c}{ROC-AUC (↑)} &
  \multicolumn{1}{c}{ROC-AUC (↑)} &
  \multicolumn{1}{c}{ROC-AUC (↑)} &
  \multicolumn{1}{c}{ROC-AUC (↑)} \\ \midrule
{\ul \textit{Specialist Models}} &
   &
   &
   &
   &
   &
   &
   &
   &
   &
   \\
InstructMol &
  NA \ &
  NA \ &
  0.0048 &
  0.0050 &
  0.0061 &
  82.1 &
  72.4 &
  NA \ &
  68.9 &
  NA \ \\
MolCA &
  $\geq$100 &
  $\geq$100 &
  $\geq$1 &
  $\geq$1 &
  $\geq$1 &
  79.8 &
  70.0 &
  89.5 &
  47.0 &
  63.0 \\
MolXPT &
  NA \ &
  NA \ &
  NA \ &
  NA \ &
  NA \ &
  88.4 &
  80.0 &
  95.3 &
  78.1 &
  71.7 \\ \midrule
{\ul \textit{Semi-Generalist Models}} &
   &
   &
   &
   &
   &
   &
   &
   &
   &
   \\
Mol-Instructions$^*$ &
  4.81 &
  $\geq$100 &
  0.0210 &
  0.0210 &
  0.0203 &
  41.7 &
  58.0 &
  47.8 &
  49.2 &
  48.2 \\
BioT5+$^*$(Cls. \& Trans.) &
  $\geq$100 &
  $\geq$100 &
  $\geq$1 &
  $\geq$1 &
  $\geq$1 &
  81.1 &
  65.1 &
  83.7 &
  67.0 &
  43.7 \\
BioT5+$^*$(Reg. \& React.) &
  $\geq$100 &
  $\geq$100 &
  0.0022 &
  0.0024 &
  0.0028 &
  65.5 &
  51.5 &
  51.0 &
  58.8 &
  52.5 \\ \midrule
{\ul \textit{Generalist Models}} &
   &
   &
   &
   &
   &
   &
   &
   &
   &
   \\
GPT-4 (5-shot) &
  1.68 &
  1.59 &
  0.0227 &
  0.0462 &
  0.0395 &
  62.5 &
  61.5 &
  51.6 &
  65.9 &
  40.5 \\
Galactica &
  4.34 &
  2.78 &
  0.2329 &
  0.0413 &
  0.2497 &
  58.4 &
  53.5 &
  78.4 &
  72.2 &
  55.9 \\
3D-MoLM$^*$ &
  3.41 &
  4.86 &
  0.0299 &
  0.0536 &
  0.0673 &
  55.5 &
  53.8 &
  53.7 &
  30.6 &
  49.7 \\
ChemDFM$^*$ &
  8.19 &
  6.21 &
  0.1204 &
  0.1262 &
  0.1694 &
  59.5 &
  50.5 &
  60.0 &
  52.4 &
  51.0 \\
LlaSMol$^*$ &
  \textbf{1.21} &
  1.01 &
  $\geq$1 &
  $\geq$1 &
  $\geq$1 &
  46.7 &
  82.4 &
  77.5 &
  70.3 &
  \textbf{78.4} \\
Mol-LLM (w/o Graph) &
  1.36 &
  0.95 &
  0.0044 &
  0.0043 &
  0.0055 &
  \textbf{80.8} &
  \textbf{84.3} &
  \textbf{85.0} &
  \textbf{76.5} &
  76.1 \\
Mol-LLM &
  1.28 &
  \textbf{0.91} &
  \textbf{0.0044} &
  \textbf{0.0043} &
  \textbf{0.0054} &
  80.5 &
  81.1 &
  82.4 &
  75.1 &
  76.3 \\ \bottomrule
\end{tabular}%
}
\label{tab:property_prediction_main}
\end{table}

\paragraph{Property Regression and Classification}
\Cref{tab:property_prediction_main} summarizes the property regression and classification results.
On most tasks, Mol-LLM outperforms every other generalist model, except for LogS, ClinTox, and SIDER.
Notably, even Mol-LLM (w/o Graph) performs on a par with the full model.
We attribute this behavior to the small molecular sizes in MoleculeNet~\cite{Wu2017MoleculeNetAB}, which allow the LLM to infer structural information directly from the SELFIES representation.

\begin{table}[t!]
\footnotesize\centering\setlength{\tabcolsep}{4.4pt}
\caption{
Performance comparison for reaction prediction tasks on Mol-Instructions~\cite{Fang2023MolInstructionsAL} and SMolInstruct \cite{Yu2024LlaSMolAL} datasets.
}
\resizebox{\columnwidth}{!} {
\begin{tabular}{@{}lcccccc@{}}
\toprule
Dataset                               & \multicolumn{6}{c}{Mol-Instructions / SMolInstruct}                                    \\ \midrule
Task                                  & \multicolumn{2}{c}{Forward Synthesis}        & \multicolumn{2}{c}{Retrosynthesis}        & \multicolumn{2}{c}{Reagent
Prediction} \\ \midrule
Metric &
  EXACT ($\uparrow$) &
  MACCS FTS ($\uparrow$) &
  EXACT ($\uparrow$) &
  MACCS FTS ($\uparrow$) &
  EXACT ($\uparrow$) &
  MACCS FTS ($\uparrow$) \\ \midrule
{\ul \textit{Specialist Models}} &
  \multicolumn{1}{r}{} &
  \multicolumn{1}{r}{} &
  \multicolumn{1}{l}{} &
  \multicolumn{1}{l}{} &
  \multicolumn{1}{l}{} &
  \multicolumn{1}{l}{} \\
InstructMol                           & 0.536 / \ \ NA \ & 0.878 / \ \ NA \ & 0.407 / \ \ \ NA \ & 0.852 / \ \ NA \ & 0.129      & 0.539     \\
MolCA$^*$                             & 0.000 / 0.000 & 0.494 / 0.357 & 0.000 / 0.000 & 0.880 / 0.760 & 0.000      & 0.115     \\ \midrule
{\ul \textit{Semi-Generalist Models}} &               &               &               &               &            &           \\
Mol-Instructions$^*$                  & 0.052 / 0.003 & 0.291 / 0.184 & 0.069 / 0.015 & 0.359 / 0.285 & 0.044      & 0.364     \\
BioT5+$^*$(Cls. \& Trans.)            & 0.000 / 0.000 & 0.152 / 0.187 & 0.001 / 0.000 & 0.195 / 0.170 & 0.000      & 0.056     \\
BioT5+$^*$(Reg. \& React.)            & 0.864 / 0.081 & 0.975 / 0.537 & 0.642 / 0.152 & 0.930 / 0.751 & 0.257      & 0.621     \\ \midrule
{\ul \textit{Generalist Models}}      &               &               &               &               &            &           \\
GPT-4 (5-shot)                        & 0.021 / 0.011 & 0.728 / 0.634 & 0.012 / 0.013 & 0.716 / 0.686 &    0.000      & 0.228         \\
Galactica                             & 0.000 / 0.000 & 0.257 / 0.377 & 0.000 / 0.000 & 0.274 / 0.447 & 0.000      & 0.127     \\
3D-MoLM$^*$                           & 0.000 / 0.000 & 0.391 / 0.296 & 0.000 / 0.000 & 0.451 / 0.372 & 0.000      & 0.218     \\
ChemDFM$^*$                           & 0.000 / 0.002 & 0.142 / 0.178 & 0.000 / 0.000 & 0.440 / 0.443 & 0.000      & 0.099     \\
LlaSMol$^*$                           & 0.743 / \textbf{0.629} & 0.955 / \textbf{0.919} & 0.453 / 0.323 & 0.885 / 0.827 & 0.000      & 0.199     \\
Mol-LLM (w/o Graph)                   & 0.893 / 0.584 & 0.983 / 0.904 & 0.510 / 0.363 & 0.886 / 0.828 & 0.202      & 0.586     \\
Mol-LLM &
  \textbf{0.911} / 0.601 &
  \textbf{0.987} / 0.908 &
  \textbf{0.538 / 0.377} &
  \textbf{0.893 / 0.832} &
  \textbf{0.225} &
  \textbf{0.600} \\ \bottomrule
\end{tabular}%
}
\label{tab:reaction_prediction_main}
\end{table}

\paragraph{Reaction Prediction}
The reaction prediction results are reported in~\Cref{tab:reaction_prediction_main}.
Except for the FS task of SMolInstruct dataset, Mol-LLM again leads all generalist models.
Since successful reaction prediction depends on recognizing which functional groups can participate during a chemical reaction, these results suggest that pre-training of the GNN on functional group prediction helps Mol-LLM exploit structural cues more effectively.
Consistent with this interpretation, omitting the graph input (w/o Graph variant) noticeably degrades performances.

\begin{table}[t!]
\footnotesize\centering\setlength{\tabcolsep}{4.4pt}
\caption{
Performance comparison for molecule generation and molecule captioning on ChEBI-20~\cite{Edwards2022TranslationBM} and SMolInstruct \cite{Yu2024LlaSMolAL} datasets.
}
\resizebox{\columnwidth}{!}{
\begin{tabular}{@{}lcccccc@{}}
\toprule
Dataset                               & \multicolumn{6}{c}{ChEBI-20 / SMolInstruct}                                                                              \\ \midrule
Task                                  & \multicolumn{3}{c}{Molecule Generation}                         & \multicolumn{3}{c}{Molecule Captioning}                \\ \midrule
Metric &
  EXACT ($\uparrow$) &
  MACCS FTS ($\uparrow$) &
  VALIDITY ($\uparrow$) &
  BLEU-4 ($\uparrow$) &
  ROUGE-L ($\uparrow$) &
  METEOR ($\uparrow$) \\ \midrule
{\ul \textit{Specialist Models}}      &                     &                     &                     &                  &                  &                  \\
GIT-Mol                               & 0.051 / \ \ NA \    & 0.738 / \ \ NA \    & 0.93 / \ \ NA \     & 0.263 / \ \ NA \ & 0.560 / \ \ NA \ & 0.533 / \ \ NA \ \\
InstructMol &
  \ \ NA \ / \ \ NA \ &
  \ \ NA \ / \ \ NA \ &
  \ \ NA \ / \ \ NA \ &
  0.371 / \ \ NA \ &
  0.502 / \ \ NA \ &
  0.509 / \ \ NA \ \\
MolT5                                 & 0.311 / 0.317       & 0.834 / 0.879       & 0.91 / 0.95         & 0.508 / 0.366    & 0.594 / 0.501    & 0.614 / 0.515    \\
MolCA$^*$                             & \ \ NA \ / \ \ NA \ & \ \ NA \ / \ \ NA \ & \ \ NA \ / \ \ NA \ & 0.540 / 0.510    & 0.631 / 0.604    & 0.652 / 0.628    \\
MolXPT                                & 0.215 / \ \ NA \    & 0.859 / \ \ NA \    & 0.98 / \ \ NA \     & 0.505 / \ \ NA \ & 0.597 / \ \ NA \ & 0.626 / \ \ NA \ \\
Text+Chem T5                          & 0.322 / \ \ NA \    & 0.901 / \ \ NA \    & 0.94 / \ \ NA \     & 0.542 / \ \ NA \ & 0.622 / \ \ NA \ & 0.648 / \ \ NA \ \\ \midrule
{\ul \textit{Semi-Generalist Models}} &                     &                     &                     &                  &                  &                  \\
Mol-Instructions                      & 0.016 / 0.045       & 0.167 / 0.475       & 1.00 / 1.00         & 0.171 / 0.020    & 0.289 / 0.217    & 0.271 / 0.124    \\
BioT5+$^*$(Cls. \& Trans.)            & 0.557 / 0.519       & 0.907 / 0.897       & 1.00 / 1.00         & 0.591 / 0.582    & 0.649 / 0.644    & 0.680 / 0.677    \\
BioT5+$^*$(Reg. \& React.)            & 0.537 / 0.416       & 0.897 / 0.867       & 1.00 / 1.00         & 0.216 / 0.221    & 0.364 / 0.364    & 0.323 / 0.321    \\ \midrule
{\ul \textit{Generalist Models}}      &                     &                     &                     &                  &                  &                  \\
GPT-4 (5-shot)                        & 0.092 / 0.027       & 0.745 / 0.726       & 0.65 / 0.74               & 0.158 / 0.125    & 0.303 / 0.273    & 0.320 / 0.274    \\
Galactica$^*$                         & 0.000 / 0.000       & 0.264 / 0.271       & 0.70 / 0.61         & 0.000 / 0.000    & 0.006 / 0.006    & 0.004 / 0.005    \\
3D-MoLM$^*$                           & 0.000 / 0.000       & 0.000 / 0.000       & 0.00 / 0.00         & 0.171 / 0.167    & 0.287 / 0.285    & 0.326 / 0.329    \\
ChemDFM$^*$                           & 0.018 / 0.041       & 0.165 / 0.297       & 0.19 / 0.13         & 0.031 / 0.035    & 0.101 / 0.108    & 0.078 / 0.085    \\
LlaSMol$^*$                           & 0.274 / 0.180       & 0.871 / 0.845       & 0.95 / 0.93         & 0.333 / 0.328    & 0.464 / 0.465    & 0.466 / 0.470    \\
Mol-LLM (w/o Graph) &
  0.431 / 0.362 &
  0.903 / \textbf{0.888} &
  \textbf{1.00} / \textbf{1.00} &
  0.482 / 0.477 &
  \textbf{0.509} / \textbf{0.490} &
  0.587 / 0.585 \\
Mol-LLM &
  \textbf{0.443 / 0.368} &
  \textbf{0.906} / 0.887 &
  \textbf{1.00} / 0.99 &
  \textbf{0.493} / \textbf{0.482} &
  0.439 / 0.433 &
  \textbf{0.599} / \textbf{0.589} \\ \bottomrule
\end{tabular}%
}
\label{tab:translation_main}
\end{table}

\paragraph{Description-guided Molecule Generation}
\Cref{tab:translation_main} shows the results for description-guided molecule generation, whose input prompts contain no molecular graphs.
Since both Mol-LLM and the w/o Graph variant receive identical inputs, their scores are nearly indistinguishable.
This confirms that Mol-LLM's ability to use graphs does not impede its instruction-following ability when graphs are absent.
On both the ChEBI-20 and SMolInstruct datasets, Mol-LLM nonetheless achieves the best results among generalist models.

\paragraph{Molecule Captioning}
As summarized in~\Cref{tab:translation_main}, Mol-LLM again surpasses all baselines.
Compared with the w/o Graph variant, the full model obtains consistently higher BLEU and METEOR scores but slightly lower ROUGE scores on both ChEBI-20 and SMolInstruct.
The pattern implies that Mol-LLM produces more concise captions: it captures the essential information while omitting peripheral details.
We believe MolPO training encourages the model to rely on structural cues and focus on the core content.

\begin{table}[t!]
\footnotesize\centering\setlength{\tabcolsep}{4.4pt}
\caption{
Evaluation of OOD generalization for reaction prediction on the ORDerly dataset, which is non-USPTO, and LogS on the AqSol dataset.
}
\resizebox{\columnwidth}{!} {
\begin{tabular}{@{}lccccccc@{}}
\toprule
Dataset                               & AqSol         & \multicolumn{6}{c}{ORDerly}                                                     \\ \midrule
Task                                  & LogS          & \multicolumn{3}{c}{Forward Synthesis}  & \multicolumn{3}{c}{Retrosynthesis}     \\ \midrule
Metric & RMSE ($\downarrow$) & EXACT ($\uparrow$) & MACCS FTS ($\uparrow$) & VALIDITY ($\uparrow$) & EXACT ($\uparrow$) & MACCS FTS ($\uparrow$) & VALIDITY ($\uparrow$) \\ \midrule
{\ul \textit{Semi-Generalist Models}} &               &                &                &      &                &                &      \\
BioT5+$^*$(Reg. \& React.)            & 1.81          & 0.095          & 0.628          & 1.00 & 0.139          & 0.678          & 1.00 \\ \midrule
{\ul \textit{Generalist Models}}      &               &                &                &      &                &                &      \\
GPT-4                                 & 2.17          & 0.000          & 0.723          & 0.87 & 0.000          & 0.672          & 0.65  \\
Galactica$^*$                         & 3.20          & 0.000          & 0.322          & 0.49 & 0.000          & 0.398          & 0.38 \\
3D-MoLM$^*$                           & 2.72          & 0.000          & 0.288          & 0.01 & 0.000          & 0.396          & 0.01 \\
ChemDFM$^*$                           & 6.98          & 0.017          & 0.428          & 0.04 & 0.000          & 0.406          & 0.05 \\
LlaSMol$^*$                           & 1.32          & 0.350          & 0.881          & \textbf{1.00} & 0.473          & 0.875          & 0.99 \\
Mol-LLM (w/o Graph)                   & 1.10          & 0.394          & \textbf{0.900} & \textbf{1.00}    & 0.727          & 0.936          & \textbf{1.00}    \\
Mol-LLM                               & \textbf{1.02} & \textbf{0.401} & 0.877          & \textbf{1.00}    & \textbf{0.738} & \textbf{0.939} & \textbf{1.00}    \\ \bottomrule
\end{tabular}%
}
\label{tab:ood_main}
\end{table}

\paragraph{Generalization Performance on Out-of-distribution Datasets}
\Cref{tab:ood_main} reports OOD results for AqSol.
On the in-distribution training tasks (LogS and SIDER), Mol-LLM lags the generalist baseline LlaSMol only marginally.
In contrast, it is markedly superior on the OOD AqSol benchmark, demonstrating stronger generalization.
A similar trend appears in the reaction prediction FS and RS tasks: Mol-LLM is slightly weaker on in-distribution FS of SMolInstruct but outperforms competitors when evaluated OOD.
These findings indicate that MolPO training confers broader generalization across both tasks and input distributions, whereas the semi-generalist BioT5+, which lacks large-scale instruction tuning, suffers a notable drop in performance.

\subsection{Ablation Study}
\label{subsec:ablation}
\paragraph{MolPO objective enhances molecular graph utilization and task performance.}
\label{subsubsec:effect_of_molpo}
To examine whether incorporating the MolPO objective $\mathcal{L}_\text{MolPO}$ during Mol-LLM training leads the model to exploit molecular graph information more effectively than training with SFT alone, we first compare, for each task $i$, the log-likelihood $r_{w,i}$ obtained when the model is given the chosen graph $g_w$ to the log-likelihood $r_{\ell,i}$ obtained when it is given the rejected graph $g_\ell$.  
We then compute the graph discrimination ratio
$
\text{GDR} = \frac{1}{N_i}\sum_{n=1}^{N_i}
    \mathbb{I}\!\big[r_{w,i}(n) > r_{\ell,i}(n)\big],
$
where $N_i$ is the number of instances in task $i$, $\mathbb{I}$ is the indicator function, and $r_{w,i}(n)$ is the log-likelihood for the $n$-th instance in task $i$.
A GDR close to 1 indicates that the model can clearly identify the correct molecular graph (i.e., it effectively exploits molecular graph information); a value near 0.5 indicates random guessing, and a value near 0 indicates systematic confusion.
\Cref{fig:generalist_graph_util} shows the per-task GDRs---green bars for MolPO-trained models and orange bars for models trained without $\mathcal{L}_\text{MolPO}$.  
The consistently higher GDRs in the MolPO setting confirm that this objective helps the model make better use of molecular graph information.
We also compare the multitask fine-tuning performance obtained when $\mathcal{L}_\text{MolPO}$ is combined with $\mathcal{L}_\text{SFT}$ to that obtained when only $\mathcal{L}_\text{SFT}$ is used.  
As shown in~\Cref{tab:effect_of_molpo_metric}, leveraging the graph modality through the MolPO objective improves performances on most tasks.

\begin{table}[t!]
\footnotesize
\centering
\caption{
An ablation study on MolPO's effect on graph utilization.
We report RMSE($\downarrow$) for LogS and LogD, and EXACT($\uparrow$) for FS, RS, RP, and T2M, each representing forward reaction prediction, retrosynthesis, reagent prediction, and molecule generation.
"Mol-Inst." and "SMol." denote the Mol-Instructions and SMolInstruct datasets, respectively.
}
\resizebox{\columnwidth}{!}{ 
\begin{tabular}{@{}lrrrrrrrrr@{}}
\toprule
 &
  \multicolumn{1}{c}{LogS} &
  \multicolumn{1}{c}{LogD} &
  \multicolumn{1}{c}{FS (Mol-Inst.)} &
  \multicolumn{1}{c}{FS (SMol.)} &
  \multicolumn{1}{c}{RS (Mol-Inst.)} &
  \multicolumn{1}{c}{RS (SMol.)} &
  \multicolumn{1}{c}{RP (Mol-Inst.)} &
  \multicolumn{1}{c}{T2M (ChEBI-20)} &
  \multicolumn{1}{c}{T2M (SMol.)} \\ \midrule
Mol-LLM (w/o MolPO) &
  1.36 &
  0.96 &
  0.907 &
  0.598 &
  0.529 &
  0.368 &
  0.220 &
  0.426 &
  0.355 \\ 
Mol-LLM &
  \textbf{1.28} &
  \textbf{0.91} &
  \textbf{0.911} &
  \textbf{0.601} &
  \textbf{0.538} &
  \textbf{0.377} &
  \textbf{0.225} &
  \textbf{0.443} &
  \textbf{0.368} \\ \bottomrule
\end{tabular}%
}
\label{tab:effect_of_molpo_metric}
\end{table}

\paragraph{Our GNN pre-training improves molecular representation.}
\label{subsubsec:effect_of_gnn_pretraining}
\begin{figure}[t!]\centering
    \includegraphics[width=.98\columnwidth]{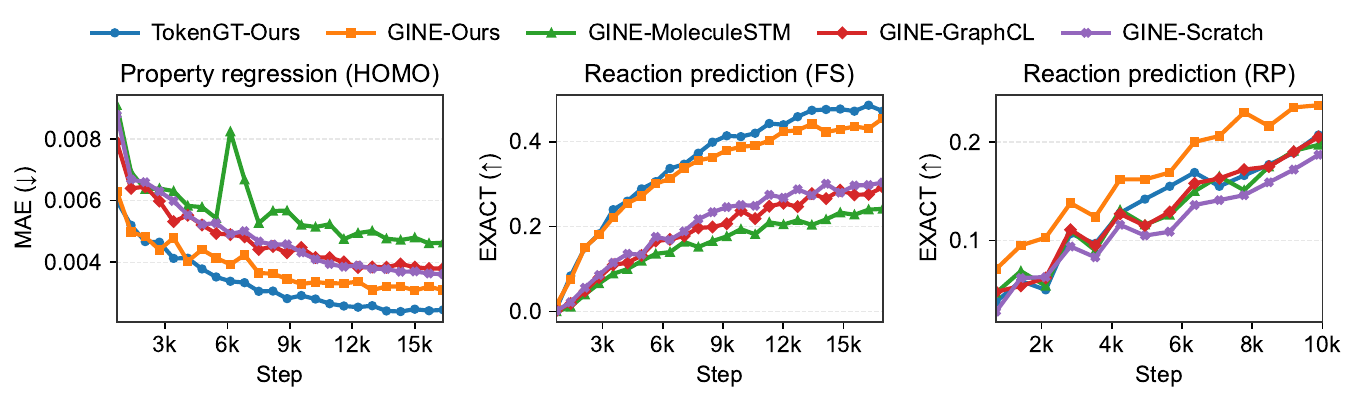}
    \caption{Comparison of fine-tuning performances on three tasks under different GNN architectures and initialization with (or without) pre-trained parameters.
Each line is labeled as \{GNN architecture\}–\{initialization\}.
\emph{Ours} refers to models initialized with parameters obtained via our GNN pre-training method, whereas \emph{Scratch} denotes models trained from random initialization. The x-axis denotes the number of training steps, and the y-axis shows the corresponding evaluation metric.}
    \label{fig:gnn_ablation}
\end{figure}
To clearly demonstrate the effect of our GNN pre-training method, we frame the experiment as a single task setting and modify Mol-LLM so that, during fine-tuning, it receives only the task instruction and the 2D molecular graph as inputs, omitting the 1D sequence.
The model is trained solely using the loss term $\mathcal{L}_{\text{SFT}}$, and its performance is then compared with different GNN architectures and weight initializations.
\Cref{fig:gnn_ablation} presents the learning curves for property regression (HOMO) and reaction prediction (FS, RP).
The GNN architectures (GINE, TokenGT) and their corresponding initializations are represented as \{GNN architecture\}-\{initialization\}.
\emph{Scratch} indicates that the GNN is trained from scratch without any pre-trained weights.
The model whose GNN is initialized with the proposed pre-training method (\emph{Ours}) consistently outperforms the others, indicating that it learns higher quality molecular representations.
Moreover, the existing pre-trained models---MoleculeSTM~\cite{liu2024multimodalmoleculestructuretextmodel} and GraphCL~\cite{you2020graph}---perform either worse than or roughly on par with the non-pretrained baseline \emph{Scratch}, which is a surprising outcome.

\section{Related Works}
\label{sec:related_works}
\paragraph{Molecular Large Language Models}
MolT5~\cite{Edwards2022TranslationBM} extends T5~\cite{raffel2023exploringlimitstransferlearning} to bidirectional translation between SMILES strings and natural language, whereas MolXPT~\cite{molxpt}, built on the GPT architecture~\cite{radford2019language}, unifies text–molecule translation with property prediction.
MolCA~\cite{Liu2023MolCAMG} and GIT-Mol~\cite{liu2024gitmol} fuse 2D molecular graphs with text via a Q-Former~\cite{li2023blip2bootstrappinglanguageimagepretraining}, while MolLM~\cite{mollm} further injects 3D geometric cues.
UniMoT~\cite{zhang2024unimotunifiedmoleculetextlanguage} discretizes Q-Former outputs into graph tokens while 3D-MolT5~\cite{pei20243dmolt5unified3dmoleculetext} introduces 3D structure tokens, enabling generative reasoning over conformers.
Although these models exploit molecule structures, each is tailored to a narrow set of tasks.
Mol-LLM tackles this limitation by jointly processing text and graphs and by performing translation, prediction, and generation within a single generalist framework.

\paragraph{Instruction-tuning on Molecular Tasks}
Mol-Instructions~\cite{Fang2023MolInstructionsAL} introduced the first broad instruction-tuning corpus, inspiring InstructMol~\cite{Cao2023InstructMolMI} to fine-tune multimodal models with task-specific prompts and BioT5+~\cite{Pei2024BioT5TG} to perform multitask tuning without structural inputs.
LlaSMol~\cite{Yu2024LlaSMolAL} scales the idea to 3.3M examples across ten tasks, yielding a single model that matches---or exceeds---specialists.
Subsequent work, including UniMoT~\cite{zhang2024unimotunifiedmoleculetextlanguage}, 3D-MolT5~\cite{pei20243dmolt5unified3dmoleculetext} and 3D-MoLM, couples instruction tuning with 2D/3D structure encoders, yet still lacks a systematic strategy for exploiting multimodal inputs.
Consequently, models remain sensitive to task distribution shifts.
Mol-LLM fills this gap by unifying instruction tuning with structure-aware training, thereby improving robustness across in-distribution and out-of-distribution tasks.

\paragraph{Preference Optimization on Different Modality}
DPO~\cite{rafailov2023dpo} aligns language models with human preferences by maximizing the log-probability gap between preferred and rejected outputs; SimPO~\cite{Meng2024SimPOSP} removes the expensive reference model for lighter training.
As multimodal LLMs rise, mDPO~\cite{Wang2024mDPOCP} adapts the idea to vision–language models by corrupting images to build preference pairs, and numerous follow-ups \cite{Xiao2024DetectingAM, Zhou2024AligningMI, Pi2024StrengtheningML, Deng2024EnhancingLV} confirm its effectiveness.
Yet no study has demonstrated comparable gains for molecular data.
Mol-LLM is the first to apply preference optimization to molecular graphs and text jointly, showing that structure-aware preferences yield stronger generalization than sequence-only tuning while keeping training costs manageable.

\section{Conclusion}
\label{sec:conclusion}
We introduced MolPO, a multimodal training objective that leverages perturbed molecules to enhance the utility of 2D molecular graphs, together with a hybrid graph encoder pre-training strategy.
We also curated a large-scale molecule instruction tuning dataset and, using the proposed methods, developed Mol-LLM, a multimodal generalist molecular large language model.
Mol-LLM achieved state-of-the-art performances among generalist molecular models on property regression, property classification, reaction prediction, description-guided molecule generation, and molecule captioning tasks.
We believe our approach can be extended beyond 2D molecular graphs to incorporate 3D structural information and molecular metadata, enabling real-world applications such as drug discovery and novel material discovery.
A detailed discussion of the limitations are described in \Cref{appx:limitation}.

\begin{ack}
LG AI Research supported this work.
This work was also supported by Artificial intelligence industrial convergence cluster development project funded by the Ministry of Science and ICT(MSIT, Korea)\&Gwangju Metropolitan City, the National Research Foundation of Korea(NRF) grant funded by the Korea government(MSIT) (No. RS-2024-00410082), Institute of Information communications Technology Planning Evaluation (IITP) grant funded by the Korea government(MSIT) (No. RS-2019-II190079, Artificial Intelligence Graduate School Program(Korea University); No.RS-2020-II201336, Artificial Intelligence Graduate School Program(UNIST); No. 2022-0-00612, Geometric and Physical Commonsense Reasoning based Behavior Intelligence for Embodied AI), and partly supported by the Institute of Information \& Communications Technology Planning \& Evaluation(IITP)-ITRC(Information Technology Research Center) grant funded by the Korea government(MSIT)(IITP-2025-RS-2024-00436857, 15\%)
\end{ack}


\bibliographystyle{unsrtnat} 
\bibliography{NeurIPS2025/reference}    

\medskip

\newpage
\appendix
\section{Limitation}
\label{appx:limitation}
\paragraph{Performance Degradation from Limited Molecular Distribution in Classification Tasks}
When the training data lacks sufficient diversity, preference optimization approaches using input preference pairs could suffer performance degradation on test or out-of-distribution datasets.
In the case of MolPO, if the training molecular distribution is too narrow or contains spurious patterns unrelated to the given molecular task, the model may inappropriately regard molecules in test set or out-of-distribution (OOD) dataset as rejected molecules, based solely on their non-in-distribution characteristics.
This hypothesis is consistent with the observations in \Cref{tab:property_prediction_main} for the classification datasets.
The classification datasets are substantially smaller than the datasets in the other task groups.
More than half of them contain only approximately 1K samples, compared with 3.3M samples in the entire training dataset, which explains why MolPO's performance either remained unchanged or slightly decreased.
The principled and necessary solution to this issue is basically to procure more diverse molecular distributions. 
We anticipate that the research community will pay more attention to developing diverse and comprehensive property classification datasets.

\paragraph{In-depth Analysis across Molecular Tasks}
Beyond the overall improvement in benchmark performance, an in-depth analysis is needed to understand what qualitative changes occur for each molecular task from the improved graph utilization by MolPO. 
It is necessary to identify trends that cannot be determined by performance metrics alone, such as which molecular features are difficult to capture with sequence-only approaches, and whether these identified molecular features have strong practical impact. 
Such analysis could be particularly interesting for property prediction tasks where spatial recognition of molecules is important.

\paragraph{Multi-step Reasoning and Multi-turn Interaction}
As demonstrated by recent successful LLMs~\citep{GPT4, Gemini}, impactful real-world applications of LLMs critically depend on multi-step reasoning capabilities and multi-turn interactions between LLMs and users. 
However, research on these two aspects remains significantly underdeveloped in the field of molecular LLMs. 
Such research requires different considerations from single-turn instruction tuning, beginning with dataset construction, and necessitates appropriate training objectives and reward modeling.
It is an interesting direction to extend molecular LLMs to multi-step reasoning and multi-turn interaction for practical applications.

\section{Implementation Details}
\label{appx:implementation_details}
This section discusses the details of the Mol-LLM implementation.
All the necessary materials to reproduce the results through \Cref{tab:property_prediction_main,tab:reaction_prediction_main,tab:translation_main,tab:ood_main}, including code, trained model, and test set, are available at \url{https://anonymous.4open.science/r/mol-llm-neurips2025-93EB}.
\begin{figure}[t!]\centering
      \begin{subfigure}[b]{0.99\columnwidth}
        \includegraphics[width=\linewidth]{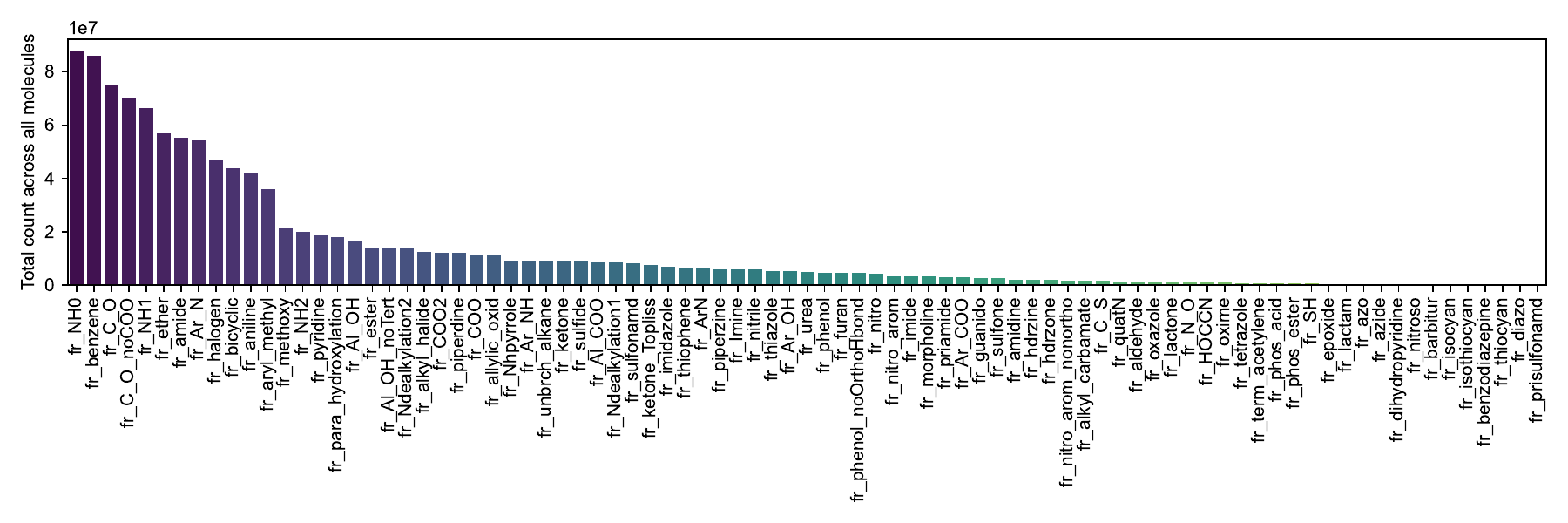}
      \end{subfigure}\hfill
      
      \begin{subfigure}[b]{0.99\columnwidth}
        \includegraphics[width=\linewidth]{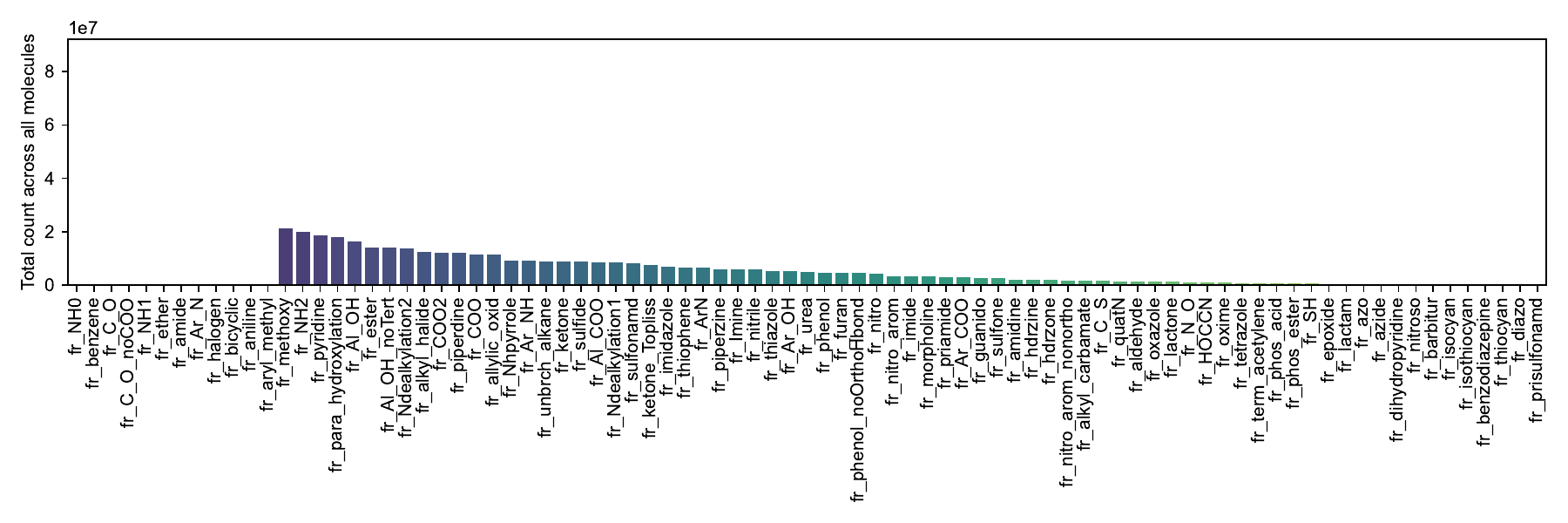}
      \end{subfigure}\hfill

     \begin{subfigure}[b]{0.99\columnwidth}
        \includegraphics[width=\linewidth]{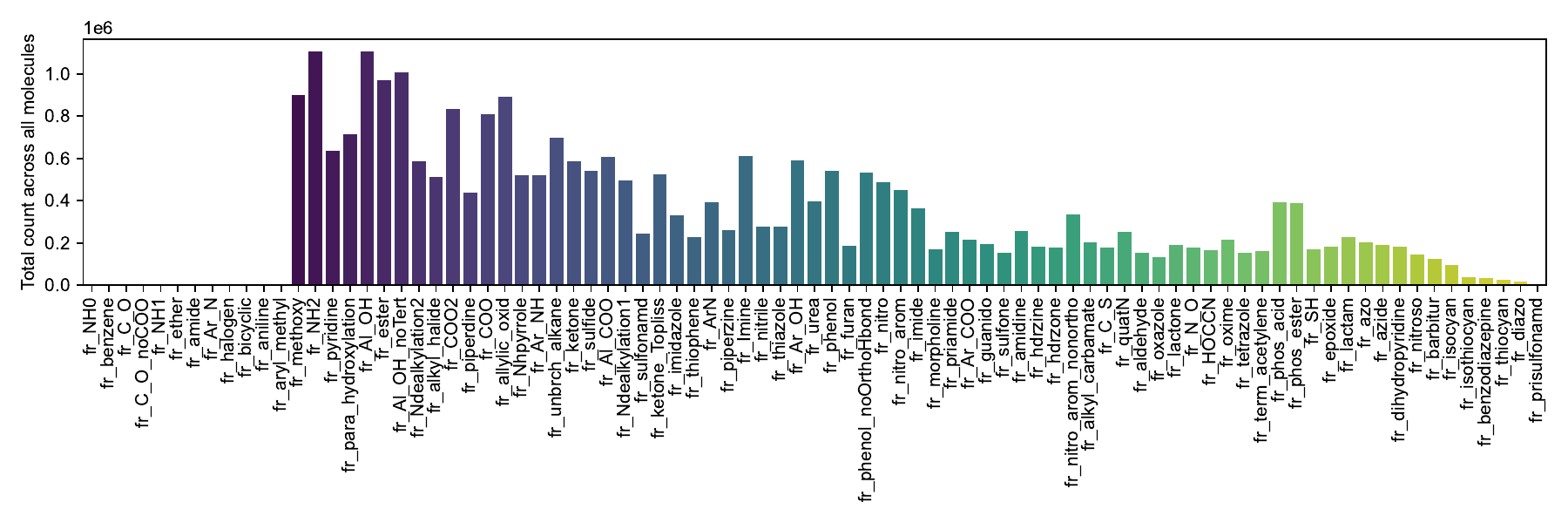}
      \end{subfigure}\hfill
    \caption{(Top) Distribution of functional groups present in molecules from the PubChem database.
(Middle) Distribution of functional groups in PubChem molecules after excluding groups that are either overly common or extremely rare.
(Bottom) Distribution of functional groups obtained after sampling 5M molecules from PubChem database, considering functional group sparsity.
Since the number of molecules differs among panels, the y-axis scale varies across plots for visualization purposes.}
    \label{fig:func_group_dist}
\end{figure}

\subsection{Functional Group Prediction Dataset for Graph Encoder Pre-training}
\label{appx:functional_group_prediction_dataset_construction}
As explained in \Cref{subsec:pre-training_stage}, the proposed graph encoder pre-training conducts functional group prediction of a given molecule, a kind of self-supervision task carried out only with the input molecule.
The principal challenge in constructing the functional group prediction dataset is the severe class imbalance: some groups occur in most molecules, whereas others are exceedingly rare.
Leveraging the RDKit Fragments module\footnote{https://www.rdkit.org/docs/source/rdkit.Chem.Fragments.html}, we enumerate 87 functional groups and quantify their occurrences across the entire PubChem database, as summarized in the top panel of~\Cref{fig:func_group_dist}.
\Cref{fig:func_group_dist} illustrates functional group imbalance, for example, \texttt{fr\_NH0} (tertiary amines) appears in many molecules, whereas \texttt{fr\_prisulfonamd} (primary sulfonamides) are scarce.
This imbalance can cause overfitting to dominant classes instead of learning general chemical knowledge.
To alleviate the overfitting problem, we remove the 11 most prevalent groups (from \texttt{fr\_NH0} to \texttt{fr\_aryl\_methyl}) and the rarest group (\texttt{fr\_prisulfonamd}), retaining $72$ functional groups.
The middle panel of~\Cref{fig:func_group_dist} shows the reduced yet still skewed distribution.
To adjust the skewed distribution, we apply sparsity-aware importance sampling as follows.
Given $M$ molecules and $G$ retained groups, let
$x_{i,g}\in\{0,1\}$ indicate the presence of group $g$ in molecule $i$.
Then we can define group frequencies as $c_g = \sum_{i=1}^{M} x_{i,g}$.
Here, we implement importance sampling that favors rarer groups by introducing the scaling factor $s_g = 1/(c_g+\varepsilon)$ with $\varepsilon=10^{-6}$, resulting in the sparsity score of molecule $i$ as
\begin{equation*}
\sigma_i=\Bigl(\sum_{g=1}^{G} x_{i,g}\,s_g\Bigr)^{2}.
\end{equation*}
Normalizing the scores yields a categorical distribution $p_i=\sigma_i\big/\sum_{j=1}^{M}\sigma_j$, from which we sample 5M molecules.
The resulting distribution (bottom panel of~\Cref{fig:func_group_dist}) is flatter than before, which enables the graph encoder to learn more unbiased chemical knowledge than when trained on the raw PubChem molecule distribution.

\subsection{Details of Graph Encoder and Pre-training}
\label{appx:details_of_graph_encoder_pre_training}
\paragraph{Architecture} Both GNN components of the hybrid graph encoder---GINE and TokenGT---use a hidden dimension \(d_g = 1024\) and five message-passing layers.
We replace the original transformer blocks of TokenGT with a BERT encoder implemented in \texttt{FlashAttention-2} and configured with eight attention heads, thereby maximizing GPU throughput.
Within TokenGT, the node- and edge-projection dimensions are both 64, and we adopt the graph laplacian eigenvector variant for node positional encoding.
\paragraph{Pre-training} GINE and TokenGT are pre-trained with the same set of hyperparameters.
For SELFIES reconstruction, sequences are truncated to a maximum length of 512 tokens; tokens beyond this limit do not contribute to the loss calculation.
The GPT-2 decoder used for reconstruction consists of six layers, eight attention heads, and an embedding size of~1024.  
We train for 50 epochs with a learning rate of \(1 \times 10^{-4}\), a batch size of~64, and the AdamW optimizer.
Training is performed on the 5M molecule dataset described in Appendix~\ref{appx:functional_group_prediction_dataset_construction}, further augmented only by adding the corresponding SELFIES strings, and all experiments are run on four NVIDIA A100 GPUs.

\begin{table}[t!]
\footnotesize\centering\setlength{\tabcolsep}{4.4pt}
\caption{
Comparison of MAE (↓) across different GNN training settings on the QM9 dataset.
}
\resizebox{\columnwidth}{!}{%
\begin{tabular}{@{}llrrr@{}}
\toprule
Property &
  Description &
  \multicolumn{1}{c}{GINE (Tuning)} &
  \multicolumn{1}{c}{GINE (Frozen)} &
  \multicolumn{1}{c}{GINE (Frozen MoleculeSTM)} \\ \midrule
$\mu$                  & Dipole moment                                     & \textbf{0.5247}    & 0.9616    & 1.0927    \\
$\alpha$               & Isotropic polarizability                          & \textbf{1.026}     & 3.1919    & 3.3589    \\
$\epsilon_\text{HOMO}$ & Highest occupied molecular orbital energy (HOMO)  & \textbf{0.1558}    & 0.2864    & 0.357     \\
$\epsilon_\text{LUMO}$ & Lowest unoccupied molecular orbital energy (LUMO) & \textbf{0.1428}    & 0.3555    & 0.4581    \\
$\Delta\epsilon$ &
  Gap between $\epsilon_\text{HOMO}$ and $\epsilon_\text{LUMO}$ (Gap) &
  \textbf{0.1817} &
  0.4003 &
  0.3997 \\
<$R^2$>                & Electronic spatial extent                         & \textbf{24.7215}   & 94.5913   & 103.2612  \\
$ZPVE$                 & Zero point vibrational energy                     & \textbf{0.0471}    & 0.3056    & 0.234     \\
$U_0$                  & Internal energy at 0K                             & \textbf{9,474.99}  & 10,302.67 & 10,176.77 \\
$U$                    & Internal energy at 298.15K                        & 10,160.12 & 10,550.07 & \textbf{10,134.84} \\
$H$                    & Enthalpy at 298.15K                               & 10,295.32 & 10,466.08 & \textbf{10,057.47} \\
$G$                    & Free energy at 298.15K                            & \textbf{9,596.59}  & 10,278.72 & 10,142.24 \\
$c_v$                  & Heat capavity at 298.15K                          & \textbf{0.5053}    & 1.1965    & 1.4149    \\
$U_0^\text{ATOM}$      & Atomization energy at 0K                          & \textbf{0.9713}    & 3.6615    & 3.2477    \\
$U^\text{ATOM}$        & Atomization energy at 298.15K                     & \textbf{0.8442}    & 3.4631    & 3.309     \\
$H^\text{ATOM}$        & Atomization enthalpy at 298.15K                   & \textbf{0.999}     & 3.6308    & 3.2287    \\
$G^\text{ATOM}$        & Atomization free energy at 298.15K                & \textbf{1.0225}    & 3.5317    & 3.4369    \\
A                      & Rotational constant                               & \textbf{0.9253}    & 0.7283    & 1.0479    \\
B                      & Rotational constant                               & \textbf{0.1515}    & 0.2428    & 0.2511    \\
C                      & Rotational constant                               & \textbf{0.0773}    & 0.172     & 0.1368    \\ \bottomrule
\end{tabular}%
}
\label{tab:molecule_rep_qm9}
\end{table}

\subsection{Investigation of Graph Encoder Used in Prior Work}
\label{appx:molecule_representation_experiment}
In \Cref{subsubsec:effect_of_gnn_pretraining}, we show that the downstream performance of the LLM integrated with pre-trained GNNs used in~\citet{Liu2023MolCAMG,Cao2023InstructMolMI}, which are MoleculeSTM~\citep{liu2024multimodalmoleculestructuretextmodel} and GraphCL~\citep{you2020graph}, does not improve from that of random initialization.
To further investigate the graph representation of MoleculeSTM, we conducted an additional experiment evaluating MoleculeSTM in isolation from the LLM on the QM9 datasets.
In this experiment, the graph embedding $h_g$ is obtained by mean-pooling the node embeddings, and then passed to a simple MLP, a regression head, whose output is used for training over MSE minimization.
While tuning the regression head, we compare three GNN tuning settings: tuning a randomly initialized GNN, freezing a randomly initialized GNN, and freezing a GNN initialized with MoleculeSTM.
\Cref{tab:molecule_rep_qm9} reports the mean absolute error (MAE) for each property.
It turns out that, when only the regression head is trained, the gap between random and MoleculeSTM initialization remains negligible w.r.t. the jointly training of GINE, reinforcing our observation that the current pre-trained GNN model fails to capture useful molecular representations.
All models were trained for 1,500 epochs with a batch size of $128$ using the Adam optimizer with a learning rate of $10^{-4}$ on four NVIDIA A100 GPUs.

\subsection{Molecular Structure Preference Pair}
\label{appx:gen_preference}
To improve the graph utilization of our model, we create molecular structural preference pairs, which are required for Molecular Structure Preference Optimization (MolPO).
Specifically, as a generalist molecular LLM, it requires a preference pair generation method applicable across various molecular tasks. 
Therefore, we employed functional group-based substructure modification, which can alter molecular features based on only the input molecule, without requiring task-specific design.
For this, we propose Molecular ACCess System (MACCS)~\citep{durant2002reoptimization} keys-based substructure modification method directly modifies molecular substructures by randomly removing and adding them.
This approach first identifies the substructures of the molecule corresponding to MACCS keys, generating two lists: one containing the MACCS keys representing functional groups present in the molecule, and the other containing the keys for functional groups absent from the molecule.
Then we sample random keys from the present MACCS keys to remove from the original molecular graph.
Subsequently, other random keys are chosen from the list of absent MACCS keys, and functional groups corresponding to the selected MACCS keys are attached at a random position in the molecule.
We set the number of MACCS keys randomly selected to 30 percent of the number of each molecule's present MACCS keys.
This method effectively alters molecular structural information without task-specific design, at the same time, it does not require heavy computation.

\subsection{Details of Mol-LLM}
\label{appx:details_of_molpo_objective}
This section describes the details of the Mol-LLM architecture and training, including the hyperparameters listed in~\Cref{tab:hyperparams_llm}.

\begin{table}[t!]
\centering
\caption{Model hyperparameters used for Mol-LLM architecture, evaluation, and training stages.}
\label{tab:hyperparams_llm}
\begin{subtable}[t]{0.43\linewidth}
    \centering
    \caption{Q-Former, LoRA, evaluation}
    \label{tab:sub_arch_eval}
\begin{tabular}{@{}lr@{}}
\toprule
Parameter                 & Value \\ \midrule
{\ul \textit{Q-Former}}   &                           \\
bert\_hidden\_dim         & 768                       \\
bert\_name                & \texttt{scibert}~\cite{scibert}            \\
num\_query\_token         & 32                        \\
bert\_layers              & 5                         \\ \midrule
{\ul \textit{LoRA}}       &                           \\
lora\_r                   & 64                        \\
lora\_alpha               & 32                        \\
lora\_dropout             & 0.1                       \\ \midrule
{\ul \textit{Evaluation}} &                           \\
gen\_max\_len             & 256                       \\
num\_beams                & 1                         \\ \bottomrule
\end{tabular}%
\end{subtable}
\hspace{1.5em}
\begin{subtable}[t]{0.51\linewidth}
    \centering
    \caption{Training hyperparameters for each stage}
    \label{tab:sub_training}
\begin{tabular}{@{}lrrr@{}}
\toprule
Parameter               & Stage 1       & Stage 2       & Stage 3               \\ \midrule
max\_length             & \multicolumn{3}{c}{512}                               \\
batch\_size             & 968           & 1024          & 1024                  \\
optimizer               & \multicolumn{3}{c}{\texttt{adamw}}                      \\
scheduler               & \multicolumn{3}{c}{\texttt{linear\_warmup\_cosine\_lr}} \\
weight\_decay           & \multicolumn{3}{c}{0.05}                              \\
min\_lr                 & \multicolumn{3}{c}{$10^{-5}$}                         \\
init\_lr                & $10^{-4}$     & $10^{-4}$     & $4 \times 10^{-5}$    \\
warmup\_lr              & $10^{-5}$     & $10^{-5}$     & $4 \times 10^{-6}$    \\
warmup\_epochs          & \multicolumn{3}{c}{0.25}                              \\
gradient\_clip\_val     & \multicolumn{3}{c}{0.5}                               \\
precision               & \multicolumn{3}{c}{\texttt{bf16-mixed}}                 \\
c                       & NA            & NA            & 0.25                  \\
$\lambda_\text{margin}$ & NA            & NA            & 0.25                  \\
$\lambda_\text{clip}$   & NA            & NA            & 1.0                   \\ \bottomrule
\end{tabular}%
\end{subtable}
\end{table}

\paragraph{Architecture}
When using Q-Former as the cross-modal projector, instead of using randomly initialized weights, we initialize it, similarly to~\citet{Liu2023MolCAMG}, using the parameters of a 12-layer pre-trained transformer encoder with an embedding dimension of 768.
However, we observed that successful multi-task learning can be achieved without fully utilizing all 12 layers of the Q-Former while maintaining performance without significant performance degradation. 
Therefore, to reduce the pre-training cost of Q-Former, we use only 5 layers instead of all 12 layers. 
The number of Q-Former query tokens is set to 32 for multi-task learning, which is more than the eight used in prior work~\cite{Liu2023MolCAMG}.
We set the LoRA rank to 64, alpha to 32, and the dropout rate to 0.1.

\paragraph{Three Stage Training}
For component ablation, we maintain identical hyperparameters for Mol-LLM, Mol-LLM (w/o Graph), and Mol-LLM (w/o MolPO), as specified in~\Cref{tab:hyperparams_llm}.
In Stage 1, along with the GNN pre-training described in~\Cref{appx:details_of_graph_encoder_pre_training}, we fine-tune only the LoRA parameters of the LLM for 12 epochs.
In Stage 2, we train the Q-Former for a single epoch to align the LLM and GNN embeddings learned in Stage 1. 
Next, in Stage 3, as described in \Cref{sec:method}, we train using the combined objective $\mathcal{L}_\text{SFT} + c\mathcal{L}_{\text{MolPO}}$, which combines both the SFT and MolPO objectives. 
Here, the scaling factor $c=0.25$ is adjusted to ensure that the scales between $\mathcal{L}_\text{SFT}$ and $c\mathcal{L}_{\text{MolPO}}$ do not differ significantly. 
For the hyperparameters used in $\mathcal{L}_{\text{MolPO}}=\mathbb{E}_{(s,q_i,g,y) \sim \mathcal{D}_\text{tr}}[-\log\sigma\bigl(\min(r_{w,i}-r_{\ell,i},\lambda_{\text{clip}}|r_{w,i}|)-\gamma_i\bigr)]$, we use $\lambda_\text{margin}=0.5$ and $\lambda_\text{clip}=1.0$, respectively.
For Stage 3, we initially trained the model for 6 epochs using the hyperparameters specified in \Cref{tab:hyperparams_llm}; however, we observed that performance had not fully converged on several tasks.
Therefore, we report experimental results based on the model trained for one additional epoch using a reduced initial learning rate of $2 \times 10^{-5}$ (half of the original value) without a warm-up epoch.

\section{Molecular Instruction-tuning Dataset}
\label{appx:dataset}
This section describes the construction details of our molecular instruction-tuning dataset, whose statistics are described in \Cref{tab:instruction_stat_full}.
It covers 21 tasks grouped into eight categories, comprising about 3.3M training and 40K test instances.

\begin{table}[t!]
\footnotesize\centering\setlength{\tabcolsep}{4.4pt}
\caption{
Details of Mol-LLM instruction-tuning training data and its sources.
}
\resizebox{\linewidth}{!}{
\begin{tabular}{@{}llrrr@{}}
\toprule
Task                                   & Data Sources          & \# Train  & \# Test & \# All    \\ \midrule
Property Prediction (Regression)       & MoleculeNet~\cite{Wu2017MoleculeNetAB}           & 359,556   & 2,519   & 362,075   \\
Property Prediction (Classification)   & MoleculeNet~\cite{Wu2017MoleculeNetAB}           & 59,607    & 7,460   & 67,067    \\
Forward Reaction Prediction            & USPTO~\cite{Wei2010ANM_uspto}                 & 1,079,379 & 5,062   & 1,084,441 \\
Retrosynthesis                         & USPTO 500MT           & 968,943   & 5,156   & 974,099   \\
Reagent Prediction                     & USPTO 500K            & 121,896   & 1,000   & 122,896   \\
Molecule Captioning                    & ChEBI-20~\cite{edwards2021text2mol} & 58,763    & 5,793   & 64,556    \\
Description-Guided Molecule Generation & ChEBI-20              & 58,763    & 5,838   & 64,601    \\
Name Conversion                        & PubChem~\cite{pubchem}               & 599,767   & -       & 599,767   \\ \midrule
Overall                                &                       & 3,306,674 & 40,757  & 3,347,431 \\ \bottomrule
\end{tabular}%
}
\label{tab:instruction_stat_full}
\vspace{-0.5cm}
\end{table}

\subsection{In-distribution Dataset Construction}
\label{appx:dataset_ind}
We integrate molecules for each task from the molecule-oriented datasets Mol-Instructions~\cite{Liu2023MolCAMG} and SMolInstruct~\cite{Yu2024LlaSMolAL}. 
During this integration process, tasks present in both datasets, such as forward synthesis and molecule captioning, are deduplicated to ensure that molecules included in the test set of one dataset do not appear in the training set of the combined dataset. 
In this process, we exclude certain tasks that are not directly relevant (e.g., NC-I2F and NC-S2F).
For tasks absent in both datasets, such as BACE, molecules are directly extracted from the original data sources to construct the dataset. 
Finally, we augment the resulting task-specific datasets with instructions using templates adopted and extended from SMolInstruct.

\subsection{Out-of-distribution Dataset Construction}
\label{appx:dataset_ood}
\paragraph{LogS - AqSol Dataset} 
To evaluate Mol-LLM on OOD LogS prediction, we use the AqSol dataset~\cite{Sorkun2019AqSolDBAC}, which contains multiple water solubility datasets in addition to ESOL.
The AqSol dataset is constructed by curating data from 9 different water solubility datasets for 9,982 unique molecules.
For our out-of-distribution evaluation on the ESOL dataset, we removed instances from the AqSol dataset that overlap with the ESOL dataset based on the molecule's InChI.
Notably, it is common for different prediction datasets to annotate different labels for the same molecule. 
This occurs due to experimental errors or when LogS labels are predicted based on different prediction models. 
To ensure high label reliability, we retain 925 molecules whose labels are either unique or have an inter-dataset standard deviation < 0.1.

\paragraph{Reaction Prediction - ORDerly Dataset} From Open Reaction Database (ORD)~\cite{Wigh2024ORDerlyDS}, we collected non-USPTO reaction data relevant to forward synthesis and retrosynthesis.
Since all reactions in our instruction-tuning dataset are derived from USPTO data, the reactions extracted from non-USPTO sources constitute out-of-distribution (OOD) samples.
Then, to ensure no duplication between the collected reaction data and those in Mol-Instructions~\cite{Fang2023MolInstructionsAL} and SMolInstruct~\cite{Yu2024LlaSMolAL}, we filtered out reactions from these non-USPTO sources whose input molecule scaffolds overlap with molecules used for reaction prediction training.
During this, we first extract data for the forward synthesis task and subsequently ensure that the retrosynthesis reaction data extraction does not duplicate entries already obtained for the forward synthesis. 
Finally, we apply scaffold splitting to each dataset, resulting in 18K training samples and 5K test samples for forward synthesis, and 54K training samples and 5K test samples for retrosynthesis.

\begin{table}[t!]
\caption{
Numbers of training and evaluation of impactful molecular tasks, which consist of property classification, property regression, reaction prediction, molecule generation, and molecule captioning, of each model. BioT5+ comprises two separate models, each trained on a distinct group of tasks.
}
\centering
\begin{tabular}{@{}lrr@{}}
\toprule
Model                                           & \# Train Tasks & \# Eval Tasks \\ \midrule
BioT5+~\cite{Pei2024BioT5TG} (Mol-Instructions) & 6                                  & 6                                 \\
BioT5+ (ChEBI-20)                               & 6                                  & 6                                 \\
LlaSMol~\cite{Yu2024LlaSMolAL}                  & 10                                 & 10                                \\
Mol-LLM (Ours)                                  & 23                                 & 15                                \\ \bottomrule
\end{tabular}%
\label{tab:train_eval_task_numbers}
\end{table}

\section{Experimental Details}
\label{appx:experimental_details}
This section provides supplementary information necessary for understanding and reproducing the main experiments. In \Cref{appx:resources}, we detail the resource requirements and execution times needed to reproduce the main results, followed by \Cref{appx:baseline_models} where we define and categorize the baseline molecular language models based on modality and task coverage. In \Cref{appx:full_exp_results}, we include full experimental results, whose evaluation metrics are skipped in the main body due to the page limit.

\subsection{Resources}
\label{appx:resources}
All experiments, except for graph encoder pre-training, were conducted on 8 NVIDIA A100 80GB GPUs and an AMD EPYC 7713 64-Core processor with 512GB of RAM.
Using this hardware configuration, Stage 1 required 6 days of training, Stage 2 required half a day, and Stage 3 required 12 days to complete.
In Stage 1 graph encoder pre-training, GINE training took approximately 18 hours on 4 A100 GPUs, and TokenGT took 19 hours.

\subsection{Baseline Models}
\label{appx:baseline_models}

\begin{table}[t!]
\caption{
Summary of baseline models categorized by their input modality and model type.
}
\label{tab:baseline_models}
\centering
\begin{tabular}{lll}
\toprule
Model         & Input Modality & Task Coverage  \\
\midrule
InstructMol~\cite{Cao2023InstructMolMI}      
    & 1D Sequence \& 2D Graph        
    & Specialist      \\
MolCA~\citep{Liu2023MolCAMG}            
    & 1D Sequence \& 2D Graph
    & Specialist      \\
MolT5~\citep{Edwards2022TranslationBM}            
    & 1D Sequence Only          
    & Specialist      \\
MolXPT~\citep{molxpt}           
    & 1D Sequence Only           
    & Specialist      \\
Mol-Instructions~\citep{Fang2023MolInstructionsAL} 
    & 1D Sequence Only           
    & Semi-Generalist \\
BioT5+~\cite{Pei2024BioT5TG}           
    & 1D Sequence Only           
    & Semi-Generalist \\
GPT-4 (5-shot)~\cite{GPT4}            
    & 1D Sequence Only           
    & Generalist      \\
Galactica~\cite{galactica}        
    & 1D Sequence Only           
    & Generalist      \\
3D-MolM~\cite{li20243dmoleculetextinterpretationlanguage}     
    & 1D Sequence \& 3D Conformer        
    & Generalist      \\
ChemDFM~\cite{zhao2024chemdfm}          
    & 1D Sequence Only           
    & Generalist      \\
LlaSMol~\cite{Yu2024LlaSMolAL}          
    & 1D Sequence Only           
    & Generalist      \\
Mol-LLM          
    & 1D Sequence \& 2D Graph
    & Generalist     \\
\bottomrule
\end{tabular}
\end{table}

As described in \Cref{subsec:experimental_setup}, we categorize the baseline models into three groups: specialist models, semi-generalist models, and generalist models, based on their level of specialization and task coverage.
In addition to the three model categories, we provide a classification based on the type of input modalities.
These categorizations are summarized in \Cref{tab:baseline_models}.

\subsubsection{Categories by Input Modalities}
\paragraph{1D Sequence Only}
Models that rely solely on 1D sequences (e.g., SMILES or SELFIES), which address molecules as strings.
This category include Galactica 6.7B~\citep{galactica}, GPT-4~\citep{GPT4}, Mol-Instructions~\citep{Fang2023MolInstructionsAL}, BioT5+\citep{Pei2024BioT5TG}, LlaSMol~\citep{Yu2024LlaSMolAL}, MolT5~\citep{Edwards2022TranslationBM}, MolXPT~\citep{molxpt}, and ChemDFM~\citep{zhao2024chemdfm}.

\paragraph{1D Sequence \& 2D Graph}
Models integrate string-based and graph-based representations to capture 2D molecular structure.
Representative examples are InstructMol~\citep{Cao2023InstructMolMI}, MolCA~\citep{Liu2023MolCAMG}, and GIT-Mol~\citep{liu2024gitmol}.
GIT-Mol additionally exploits molecular images, providing another route to leverage structural information.

\paragraph{1D Sequence \& 3D Conformer}
Models incorporate 3D conformers alongside sequence information to enrich molecular 3D spatial representations
3D-MoLM~\citep{li20243dmoleculetextinterpretationlanguage} belongs to this category.

\subsubsection{Categories by Task Coverage}
As described in \Cref{subsec:experimental_setup}, the baseline models are categorized as follows:

\paragraph{Specialist Models}
MolCA~\citep{Liu2023MolCAMG}, InstructMol~\citep{Cao2023InstructMolMI}, MolXPT~\citep{molxpt}, GIT-Mol~\citep{liu2024gitmol}, and MolT5~\citep{Edwards2022TranslationBM} are optimized for individual molecular tasks without parameter or knowledge sharing across tasks. 

\paragraph{Semi-Generalist Models}
BioT5+~\citep{Pei2024BioT5TG} and Mol-Instructions~\citep{Fang2023MolInstructionsAL} address related task groups within a single framework.
For instance, BioT5+ trains two separate models: one for classification and translation, and the other for regression and reaction prediction, enabling knowledge sharing within each group while preserving task-specific optimization.

\paragraph{Generalist Models}
Galactica 6.7B~\citep{galactica}, GPT-4~\citep{GPT4}, LlaSMol~\citep{Yu2024LlaSMolAL}, and ChemDFM~\citep{zhao2024chemdfm} aim for broad generalization by simultaneously tackling all molecular task groups.

\subsection{Full Experimental Results}
\label{appx:full_exp_results}
\begin{figure}[t!]\centering
    \includegraphics[width=.98\columnwidth]{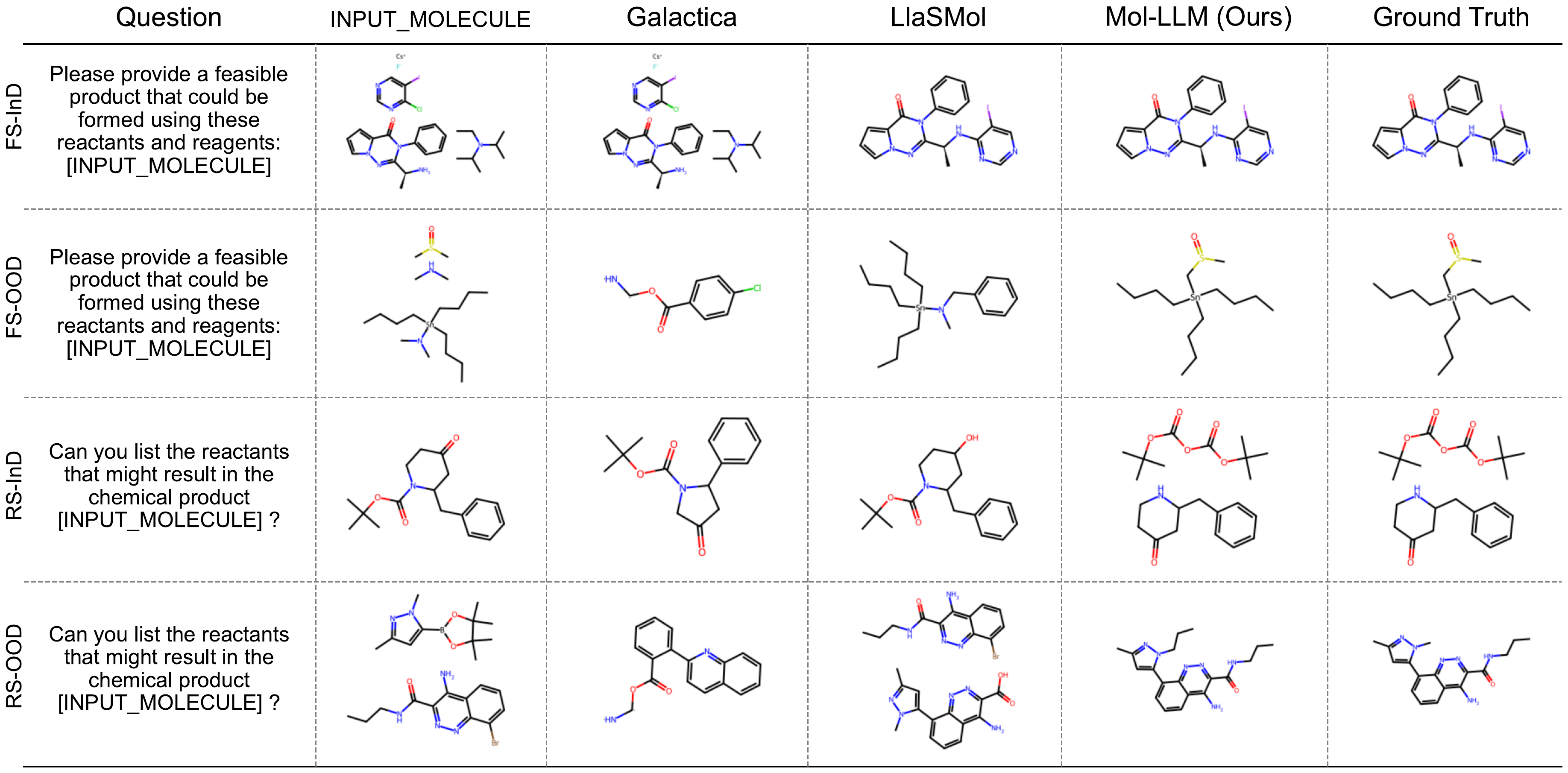}
    \caption{Comparison of predicted outputs by generalists on forward synthesis (FS) and retrosynthesis (RS), both in Mol-Instructions and ORDerly dataset.
    The upper two rows represent forward synthesis in Mol-Instructions (InD) and ORDerly (OOD) Datasets, respectively, and the lower two rows represent the retrosynthesis task in the same dataset order.    }
    \label{fig:sup_case_study}
\end{figure}

\Cref{tab:reaction_prediction_full} presents the complete results corresponding to \Cref{tab:reaction_prediction_main}.  
\Cref{tab:mol2text_full} and \Cref{tab:text2mol_full} show the full results for \Cref{tab:translation_main}.
In \Cref{fig:sup_case_study}, we also visualize predicted outputs by generalists, including Mol-LLM, Galactica~\citep{galactica}, and LlaSMol~\citep{Yu2024LlaSMolAL} on forward reaction prediction and retrosynthesis on both Mol-Instructions and ORDerly datasets.

\begin{table}[t!]
\footnotesize\centering\setlength{\tabcolsep}{4.4pt}
\caption{
Performance comparison on reaction prediction task on Mol-Instructions~\cite{Fang2023MolInstructionsAL} and SMolInstruct \cite{Yu2024LlaSMolAL} datasets.
FS, RS, RP each represent Forward synthesis, Retrosynthesis, and Reagent prediction.
}
\resizebox{\columnwidth}{!} {
\begin{tabular}{@{}lllrrrrrr@{}}
\toprule
Task &
  Dataset &
  Model &
  \multicolumn{1}{l}{EXACT ($\uparrow$)} &
  \multicolumn{1}{l}{BLEU ($\uparrow$)} &
  \multicolumn{1}{l}{RDK FTS ($\uparrow$)} &
  \multicolumn{1}{l}{MACCS FTS ($\uparrow$)} &
  \multicolumn{1}{l}{MORGAN FTS ($\uparrow$)} &
  \multicolumn{1}{l}{VALIDITY ($\uparrow$)} \\ \midrule
\multirow{31}{*}{FS} &
  \multirow{16}{*}{Mol-Instructions} &
  {\ul \textit{Specialist Models}} &
   &
   &
   &
   &
   &
   \\
 &
   &
  InstructMol &
  0.536 &
  0.967 &
  0.776 &
  0.878 &
  0.741 &
  1.00 \\
 &
   &
  MolCA$^*$ &
  0.000 &
  0.321 &
  0.329 &
  0.494 &
  0.253 &
  0.01 \\ \cmidrule(l){3-9} 
 &
   &
  {\ul \textit{Semi-Generalist Models}} &
   &
   &
   &
   &
   &
   \\
 &
   &
  Mol-Instructions$^*$ &
  0.052 &
  0.302 &
  0.232 &
  0.291 &
  0.197 &
  1.00 \\
 &
   &
  BioT5+$^*$(Cls. \& Trans.) &
  0.000 &
  0.206 &
  0.081 &
  0.152 &
  0.069 &
  0.98 \\
 &
   &
  BioT5+$^*$(Reg. \& React.) &
  0.864 &
  0.993 &
  0.949 &
  0.975 &
  0.935 &
  1.00 \\ \cmidrule(l){3-9} 
 &
   &
  {\ul \textit{Generalist Models}} &
   &
   &
   &
   &
   &
   \\
 &
   &
  GPT-4 (5-shot) &
  0.021 &
  0.580 &
  0.627 &
  0.728 &
  0.557 &
  0.93 \\
 &
   &
  Galactica &
  0.000 &
  0.468 &
  0.156 &
  0.257 &
  0.097 &
  0.95 \\
 &
   &
  3D-MoLM$^*$ &
  0.000 &
  0.081 &
  0.223 &
  0.391 &
  0.098 &
  0.01 \\
 &
   &
  ChemDFM$^*$ &
  0.000 &
  0.028 &
  0.104 &
  0.142 &
  0.077 &
  0.07 \\
 &
   &
  LlaSMol$^*$ &
  0.743 &
  0.835 &
  0.920 &
  0.955 &
  0.910 &
  0.95 \\
 &
   &
  Mol-LLM (w/o Graph) &
  0.893 &
  0.963 &
  0.968 &
  0.983 &
  0.960 &
  \textbf{1.00} \\
 &
   &
  Mol-LLM &
  \textbf{0.911} &
  \textbf{0.969} &
  \textbf{0.976} &
  \textbf{0.987} &
  \textbf{0.967} &
  \textbf{1.00} \\ \cmidrule(l){2-9} 
 &
  \multirow{15}{*}{SMolInstruct} &
  {\ul \textit{Specialist Models}} &
   &
   &
   &
   &
   &
   \\
 &
   &
  MolCA$^*$ &
  0.000 &
  0.209 &
  0.252 &
  0.357 &
  0.196 &
  0.01 \\ \cmidrule(l){3-9} 
 &
   &
  {\ul \textit{Semi-Generalist Models}} &
   &
   &
   &
   &
   &
   \\
 &
   &
  Mol-Instructions$^*$ &
  0.003 &
  0.149 &
  0.139 &
  0.184 &
  0.111 &
  1.00 \\
 &
   &
  BioT5+$^*$(Cls. \& Trans.) &
  0.000 &
  0.286 &
  0.107 &
  0.187 &
  0.089 &
  0.97 \\
 &
   &
  BioT5+$^*$(Reg. \& React.) &
  0.081 &
  0.455 &
  0.418 &
  0.537 &
  0.376 &
  1.00 \\ \cmidrule(l){3-9} 
 &
   &
  {\ul \textit{Generalist Models}} &
   &
   &
   &
   &
   &
   \\
 &
   &
  GPT-4 (5-shot) &
  0.011 &
  0.451 &
  0.520 &
  0.634 &
  0.440 &
  0.87 \\
 &
   &
  Galactica$^*$ &
  0.000 &
  0.241 &
  0.292 &
  0.377 &
  0.202 &
  0.36 \\
 &
   &
  3D-MoLM$^*$ &
  0.000 &
  0.086 &
  0.226 &
  0.296 &
  0.117 &
  0.01 \\
 &
   &
  ChemDFM$^*$ &
  0.002 &
  0.046 &
  0.125 &
  0.178 &
  0.109 &
  0.08 \\
 &
   &
  LlaSMol$^*$ &
  \textbf{0.629} &
  \textbf{0.883} &
  \textbf{0.871} &
  \textbf{0.919} &
  \textbf{0.848} &
  0.99 \\
 &
   &
  Mol-LLM (w/o Graph) &
  0.584 &
  0.867 &
  0.847 &
  0.904 &
  0.815 &
  \textbf{1.00} \\
 &
   &
  Mol-LLM &
  0.601 &
  0.873 &
  0.853 &
  0.908 &
  0.823 &
  \textbf{1.00} \\ \midrule
\multirow{31}{*}{RS} &
  \multirow{16}{*}{Mol-Instructions} &
  {\ul \textit{Specialist Models}} &
   &
   &
   &
   &
   &
   \\
 &
   &
  InstructMol &
  0.407 &
  0.941 &
  0.753 &
  0.852 &
  0.714 &
  1.00 \\
 &
   &
  MolCA$^*$ &
  0.000 &
  0.652 &
  0.936 &
  0.880 &
  0.722 &
  0.01 \\ \cmidrule(l){3-9} 
 &
   &
  {\ul \textit{Semi-Generalist Models}} &
   &
   &
   &
   &
   &
   \\
 &
   &
  Mol-Instructions$^*$ &
  0.069 &
  0.407 &
  0.303 &
  0.359 &
  0.268 &
  1.00 \\
 &
   &
  BioT5+$^*$(Cls. \& Trans.) &
  0.001 &
  0.095 &
  0.114 &
  0.195 &
  0.104 &
  0.97 \\
 &
   &
  BioT5+$^*$(Reg. \& React.) &
  0.642 &
  0.969 &
  0.897 &
  0.930 &
  0.866 &
  1.00 \\ \cmidrule(l){3-9} 
 &
   &
  {\ul \textit{Generalist Models}} &
   &
   &
   &
   &
   &
   \\
 &
   &
  GPT-4 (5-shot) &
  0.012 &
  0.573 &
  0.531 &
  0.716 &
  0.506 &
  0.77 \\
 &
   &
  Galactica &
  0.000 &
  0.452 &
  0.167 &
  0.274 &
  0.134 &
  0.99 \\
 &
   &
  3D-MoLM$^*$ &
  0.000 &
  0.069 &
  0.270 &
  0.451 &
  0.117 &
  0.01 \\
 &
   &
  ChemDFM$^*$ &
  0.000 &
  0.224 &
  0.360 &
  0.440 &
  0.234 &
  0.03 \\
 &
   &
  LlaSMol$^*$ &
  0.453 &
  0.722 &
  0.826 &
  0.885 &
  0.788 &
  0.95 \\
 &
   &
  Mol-LLM (w/o Graph) &
  0.510 &
  0.839 &
  0.835 &
  0.886 &
  0.797 &
  \textbf{1.00} \\
 &
   &
  Mol-LLM &
  \textbf{0.538} &
  \textbf{0.845} &
  \textbf{0.843} &
  \textbf{0.893} &
  \textbf{0.808} &
  \textbf{1.00} \\ \cmidrule(l){2-9} 
 &
  \multirow{15}{*}{SMolInstruct} &
  {\ul \textit{Specialist Models}} &
   &
   &
   &
   &
   &
   \\
 &
   &
  MolCA$^*$ &
  0.000 &
  0.503 &
  0.716 &
  0.760 &
  0.589 &
  0.01 \\ \cmidrule(l){3-9} 
 &
   &
  {\ul \textit{Semi-Generalist Models}} &
   &
   &
   &
   &
   &
   \\
 &
   &
  Mol-Instructions$^*$ &
  0.015 &
  0.402 &
  0.223 &
  0.285 &
  0.191 &
  1.00 \\
 &
   &
  BioT5+$^*$(Cls. \& Trans.) &
  0.000 &
  0.085 &
  0.095 &
  0.170 &
  0.085 &
  0.97 \\
 &
   &
  BioT5+$^*$(Reg. \& React.) &
  0.152 &
  0.662 &
  0.623 &
  0.751 &
  0.567 &
  1.00 \\ \cmidrule(l){3-9} 
 &
   &
  {\ul \textit{Generalist Models}} &
   &
   &
   &
   &
   &
   \\
 &
   &
  GPT-4 (5-shot) &
  0.013 &
  0.523 &
  0.499 &
  0.686 &
  0.465 &
  0.76 \\
 &
   &
  Galactica$^*$ &
  0.000 &
  0.346 &
  0.341 &
  0.447 &
  0.272 &
  0.43 \\
 &
   &
  3D-MoLM$^*$ &
  0.000 &
  0.162 &
  0.220 &
  0.372 &
  0.128 &
  0.01 \\
 &
   &
  ChemDFM$^*$ &
  0.000 &
  0.257 &
  0.304 &
  0.443 &
  0.252 &
  0.03 \\
 &
   &
  LlaSMol$^*$ &
  0.323 &
  0.759 &
  0.749 &
  0.827 &
  0.699 &
  0.99 \\
 &
   &
  Mol-LLM (w/o Graph) &
  0.363 &
  0.772 &
  0.752 &
  0.828 &
  0.699 &
  \textbf{1.00} \\
 &
   &
  Mol-LLM &
  \textbf{0.377} &
  \textbf{0.779} &
  \textbf{0.760} &
  \textbf{0.832} &
  \textbf{0.707} &
  \textbf{1.00} \\ \midrule
\multirow{16}{*}{RP} &
  \multirow{16}{*}{Mol-Instructions} &
  {\ul \textit{Specialist Models}} &
   &
   &
   &
   &
   &
   \\
 &
   &
  InstructMol &
  0.129 &
  0.610 &
  0.444 &
  0.539 &
  0.400 &
  1.00 \\
 &
   &
  MolCA$^*$ &
  0.000 &
  0.002 &
  0.033 &
  0.115 &
  0.012 &
  0.01 \\ \cmidrule(l){3-9} 
 &
   &
  {\ul \textit{Semi-Generalist Models}} &
   &
   &
   &
   &
   &
   \\
 &
   &
  Mol-Instructions &
  0.044 &
  0.224 &
  0.237 &
  0.364 &
  0.213 &
  1.00 \\
 &
   &
  BioT5+$^*$(Cls. \& Trans.) &
  0.000 &
  0.169 &
  0.038 &
  0.056 &
  0.015 &
  0.96 \\
 &
   &
  BioT5+$^*$(Reg. \& React.) &
  0.257 &
  0.695 &
  0.539 &
  0.621 &
  0.512 &
  1.00 \\ \cmidrule(l){3-9} 
 &
   &
  {\ul \textit{Generalist Models}} &
   &
   &
   &
   &
   &
   \\
 &
   &
  GPT-4 (5-shot) &
  0.000 &
  0.133 &
  0.077 &
  0.228 &
  0.071 &
  0.72 \\
 &
   &
  Galactica &
  0.000 &
  0.141 &
  0.036 &
  0.127 &
  0.051 &
  0.99 \\
 &
   &
  3D-MoLM$^*$ &
  0.000 &
  0.042 &
  0.039 &
  0.218 &
  0.077 &
  0.01 \\
 &
   &
  ChemDFM$^*$ &
  0.000 &
  0.014 &
  0.033 &
  0.099 &
  0.027 &
  0.06 \\
 &
   &
  LlaSMol$^{*}$ &
  0.000 &
  0.050 &
  0.041 &
  0.199 &
  0.050 &
  0.93 \\
 &
   &
  Mol-LLM (w/o Graph) &
  0.202 &
  0.557 &
  0.497 &
  0.586 &
  0.461 &
  \textbf{1.00} \\
 &
   &
  Mol-LLM &
  \textbf{0.225} &
  \textbf{0.578} &
  \textbf{0.517} &
  \textbf{0.600} &
  \textbf{0.485} &
  \textbf{1.00} \\ \bottomrule
\end{tabular}%
}
\label{tab:reaction_prediction_full}
\end{table}
\begin{table}[t]
\footnotesize\centering\setlength{\tabcolsep}{4.4pt}
\caption{
Performance comparison on description-guided molecule generation task on ChEBI-20~\cite{Edwards2022TranslationBM} and SMolInstruct \cite{Yu2024LlaSMolAL} datasets.
}
\resizebox{\columnwidth}{!}{
\begin{tabular}{@{}llrrrrrr@{}}
\toprule
Dataset &
  Model &
  \multicolumn{1}{l}{EXACT ($\uparrow$)} &
  \multicolumn{1}{l}{BLEU ($\uparrow$)} &
  \multicolumn{1}{l}{RDK FTS ($\uparrow$)} &
  \multicolumn{1}{l}{MACCS FTS ($\uparrow$)} &
  \multicolumn{1}{l}{MORGAN FTS ($\uparrow$)} &
  \multicolumn{1}{l}{VALIDITY ($\uparrow$)} \\ \midrule
\multirow{18}{*}{ChEBI-20}     & {\ul \textit{Specialist Models}}      &                &                &                &                &                &      \\
                               & GIT-Mol                               & 0.051          & 0.756          & 0.582          & 0.738          & 0.519          & 0.93 \\
                               & MolT5                                 & 0.311          & 0.854          & 0.746          & 0.834          & 0.684          & 0.91 \\
                               & MolXPT                                & 0.215          & NA             & 0.757          & 0.859          & 0.667          & 0.98 \\
                               & Text+Chem T5                          & 0.322          & 0.853          & 0.816          & 0.901          & 0.757          & 0.94 \\ \cmidrule(l){2-8} 
                               & {\ul \textit{Semi-Generalist Models}} &                &                &                &                &                &      \\
                               & Mol-Instructions$^*$                  & 0.016          & 0.042          & 0.132          & 0.167          & 0.090          & 1.00 \\
                               & BioT5+$^*$(Cls. \& Trans.)            & 0.557          & 0.931          & 0.835          & 0.907          & 0.780          & 1.00 \\
                               & BioT5+$^*$(Reg. \& React.)            & 0.537          & 0.821          & 0.831          & 0.897          & 0.773          & 1.00 \\ \cmidrule(l){2-8} 
                               & {\ul \textit{Generalist Models}}      &                &                &                &                &                &      \\
                               & GPT-4 (5-shot)                        & 0.092        & 0.485           & 0.518           & 0.745           & 0.482        & 0.65  \\
                               & Galactica$^*$                         & 0.000          & 0.189          & 0.142          & 0.264          & 0.057          & 0.70 \\
                               & 3D-MoLM$^*$                           & 0.000          & 0.000          & 0.000          & 0.000          & 0.000          & 0.00 \\
                               & ChemDFM$^*$                            & 0.018          & 0.205          & 0.136          & 0.165          & 0.110          & 0.19 \\
                               & LlaSMol$^*$                           & 0.274          & 0.644          & 0.755          & 0.871          & 0.679          & 0.95 \\
                               & Mol-LLM (w/o Graph)                   & 0.431          & 0.792          & 0.823          & 0.903          & 0.754          & \textbf{1.00} \\
                               & Mol-LLM                               & \textbf{0.443} & \textbf{0.795}          & \textbf{0.829} & \textbf{0.906} & \textbf{0.761} & \textbf{1.00} \\ \midrule
\multirow{15}{*}{SMolInstruct} & {\ul \textit{Specialist Models}}      &                &                &                &                &                &      \\
                               & MolT5                                 & 0.317          & NA              & 0.802          & 0.879          & 0.732          & 0.95 \\ \cmidrule(l){2-8} 
                               & {\ul \textit{Semi-Generalist Models}} &                &                &                &                &                &      \\
                               & Mol-Instructions$^*$                  & 0.045          & 0.507          & 0.366          & 0.475          & 0.272          & 1.00 \\
                               & BioT5+$^*$(Cls. \& Trans.)            & 0.519          & 0.918          & 0.822          & 0.897          & 0.757          & 1.00 \\
                               & BioT5+$^*$(Reg. \& React.)            & 0.416          & 0.819          & 0.782          & 0.867          & 0.706          & 1.00 \\ \cmidrule(l){2-8} 
                               & {\ul \textit{Generalist Models}}      &                &                &                &                &                &      \\
                               & GPT-4 (5-shot)                        & 0.027          & 0.404           & 0.482          & 0.726          & 0.368          & 0.74 \\
                               & Galactica$^*$                         & 0.000          & 0.173          & 0.144          & 0.271          & 0.055          & 0.61 \\
                               & 3D-MoLM$^*$                           & 0.000          & 0.000          & 0.000          & 0.000          & 0.000          & 0.00 \\
                               & ChemDFM$^*$                           & 0.041          & 0.069          & 0.230          & 0.297          & 0.189          & 0.13 \\
                               & LlaSMol$^*$                           & 0.180          & 0.718          & 0.712          & 0.845          & 0.623          & 0.93 \\
                               & Mol-LLM (w/o Graph)                   & 0.362          & 0.759          & 0.797          & \textbf{0.888} & 0.716          & \textbf{1.00} \\
                               & Mol-LLM                               & \textbf{0.368} & \textbf{0.761} & \textbf{0.800} & 0.887          & \textbf{0.721} & 0.99 \\ \bottomrule
\end{tabular}%
}
\label{tab:text2mol_full}
\end{table}
\begin{table}[t]
\footnotesize\centering\setlength{\tabcolsep}{4.4pt}
\caption{
Performance comparison on molecule captioning task on ChEBI-20~\cite{Edwards2022TranslationBM} and SMolInstruct \cite{Yu2024LlaSMolAL} datasets.
}
\resizebox{\columnwidth}{!}{
\begin{tabular}{@{}llrrrrrr@{}}
\toprule
 &
  Model &
  \multicolumn{1}{l}{BLEU-2 ($\uparrow$)} &
  \multicolumn{1}{l}{BLEU-4 ($\uparrow$)} &
  \multicolumn{1}{l}{ROUGE-1 ($\uparrow$)} &
  \multicolumn{1}{l}{ROGUE-2 ($\uparrow$)} &
  \multicolumn{1}{l}{ROUGE-L ($\uparrow$)} &
  \multicolumn{1}{l}{METEOR ($\uparrow$)} \\ \midrule
\multirow{20}{*}{ChEBI-20} &
  {\ul \textit{Specialist Models}} &
   &
   &
   &
   &
   &
   \\
 &
  GIT-Mol &
  0.352 &
  0.263 &
  0.575 &
  0.485 &
  0.560 &
  0.533 \\
 &
  InstructMol &
  0.475 &
  0.371 &
  0.566 &
  0.394 &
  0.502 &
  0.509 \\
 &
  MolT5 &
  0.594 &
  0.508 &
  0.654 &
  0.510 &
  0.594 &
  0.614 \\
 &
  MolCA$^*$ &
  0.623 &
  0.540 &
  0.693 &
  0.553 &
  0.631 &
  0.652 \\
 &
  MolXPT &
  0.594 &
  0.505 &
  0.660 &
  0.511 &
  0.597 &
  0.626 \\
 &
  Text+Chem T5 &
  0.625 &
  0.542 &
  0.682 &
  0.543 &
  0.622 &
  0.648 \\ \cmidrule(l){2-8} 
 &
  {\ul \textit{Semi-Generalist Models}} &
   &
   &
   &
   &
   &
   \\
 &
  Mol-Instructions &
  0.249 &
  0.171 &
  0.331 &
  0.206 &
  0.289 &
  0.271 \\
 &
  BioT5+$^*$(Cls. \& Trans.) &
  0.666 &
  0.591 &
  0.709 &
  0.583 &
  0.649 &
  0.680 \\
 &
  BioT5+$^*$(Reg. \& React.) &
  0.249 &
  0.216 &
  0.387 &
  0.302 &
  0.364 &
  0.323 \\ \cmidrule(l){2-8} 
 &
  {\ul \textit{Generalist Models}} &
   &
   &
   &
   &
   &
   \\
 &
  GPT-4 (5-shot) &
  0.261 &
  0.158 &
  0.286 &
  0.188 &
  0.303 &
  0.320 \\
 &
  Galactica$^*$ &
  0.001 &
  0.000 &
  0.006 &
  0.000 &
  0.006 &
  0.004 \\
 &
  3D-MoLM$^*$ &
  0.252 &
  0.171 &
  0.361 &
  0.184 &
  0.287 &
  0.326 \\
 &
  ChemDFM$^*$ &
  0.054 &
  0.031 &
  0.120 &
  0.049 &
  0.101 &
  0.078 \\
 &
  LlaSMol$^*$ &
  0.432 &
  0.333 &
  0.522 &
  0.356 &
  0.464 &
  0.466 \\
 &
  Mol-LLM (w/o Graph) &
  0.556 &
  0.482 &
  \textbf{0.565} &
  \textbf{0.417} &
  \textbf{0.509} &
  0.587 \\
 &
  Mol-LLM &
  \textbf{0.566} &
  \textbf{0.493} &
  0.493 &
  0.336 &
  0.439 &
  \textbf{0.599} \\ \midrule
\multirow{16}{*}{SMolInstruct} &
  {\ul \textit{Specialist Models}} &
   &
   &
   &
   &
   &
   \\
 &
  MolT5 &
  0.462 &
  0.366 &
  0.563 &
  0.398 &
  0.501 &
  0.515 \\
 &
  MolCA$^*$ &
  0.599 &
  0.510 &
  0.665 &
  0.519 &
  0.604 &
  0.628 \\ \cmidrule(l){2-8} 
 &
  {\ul \textit{Semi-Generalist Models}} &
   &
   &
   &
   &
   &
   \\
 &
  Mol-Instructions &
  0.028 &
  0.020 &
  0.226 &
  0.160 &
  0.217 &
  0.124 \\
 &
  BioT5+$^*$(Cls. \& Trans.) &
  0.656 &
  0.582 &
  0.702 &
  0.576 &
  0.644 &
  0.677 \\
 &
  BioT5+$^*$(Reg. \& React.) &
  0.257 &
  0.221 &
  0.387 &
  0.301 &
  0.364 &
  0.321 \\ \cmidrule(l){2-8} 
 &
  {\ul \textit{Generalist Models}} &
   &
   &
   &
   &
   &
   \\
 &
  GPT-4 (5-shot) &
  0.220 &
  0.125 &
  0.352 &
  0.156 &
  0.273 &
  0.274 \\
 &
  Galactica$^*$ &
  0.002 &
  0.000 &
  0.007 &
  0.000 &
  0.006 &
  0.005 \\
 &
  3D-MoLM$^*$ &
  0.244 &
  0.167 &
  0.357 &
  0.185 &
  0.285 &
  0.329 \\
 &
  ChemDFM$^*$ &
  0.057 &
  0.035 &
  0.128 &
  0.054 &
  0.108 &
  0.085 \\
 &
  LlaSMol$^*$ &
  0.427 &
  0.328 &
  0.525 &
  0.359 &
  0.465 &
  0.470 \\
 &
  Mol-LLM (w/o Graph) &
  0.554 &
  0.477 &
  \textbf{0.544} &
  \textbf{0.393} &
  \textbf{0.490} &
  0.585 \\
 &
  Mol-LLM &
  \textbf{0.558} &
  \textbf{0.482} &
  0.485 &
  0.330 &
  0.433 &
  \textbf{0.589} \\ \bottomrule
\end{tabular}%
}
\label{tab:mol2text_full}
\end{table}

\section{Broader Impacts}
\label{appx:broader_impacts}
We currently anticipate no major negative social impacts from this research; nevertheless, there is a possibility that it could be used to generate molecules harmful to humans or the environment.
At present, training is carried out on eight NVIDIA A100 GPUs, but scaling to larger LLMs would require additional GPUs and would therefore increase carbon emissions.
On the positive side, Mol-LLM enables researchers performing chemical experiments to predict experimental outcomes in advance.
\FloatBarrier 
\section*{NeurIPS Paper Checklist}

\begin{enumerate}

\item {\bf Claims}
    \item[] Question: Do the main claims made in the abstract and introduction accurately reflect the paper's contributions and scope?
    \item[] Answer: \answerYes{} 
    \item[] Justification: The claims have been validated through extensive evaluation on various molecular tasks in \Cref{sec:experiments} and \Cref{appx:full_exp_results}, comparing with a number of generalist baselines available, including in-domain and out-of-domain benchmarks.
    \item[] Guidelines:
    \begin{itemize}
        \item The answer NA means that the abstract and introduction do not include the claims made in the paper.
        \item The abstract and/or introduction should clearly state the claims made, including the contributions made in the paper and important assumptions and limitations. A No or NA answer to this question will not be perceived well by the reviewers. 
        \item The claims made should match theoretical and experimental results, and reflect how much the results can be expected to generalize to other settings. 
        \item It is fine to include aspirational goals as motivation as long as it is clear that these goals are not attained by the paper. 
    \end{itemize}

\item {\bf Limitations}
    \item[] Question: Does the paper discuss the limitations of the work performed by the authors?
    \item[] Answer: \answerYes{} 
    \item[] Justification: We discuss limitations in \Cref{sec:conclusion,appx:limitation}.
    \item[] Guidelines:
    \begin{itemize}
        \item The answer NA means that the paper has no limitation while the answer No means that the paper has limitations, but those are not discussed in the paper. 
        \item The authors are encouraged to create a separate "Limitations" section in their paper.
        \item The paper should point out any strong assumptions and how robust the results are to violations of these assumptions (e.g., independence assumptions, noiseless settings, model well-specification, asymptotic approximations only holding locally). The authors should reflect on how these assumptions might be violated in practice and what the implications would be.
        \item The authors should reflect on the scope of the claims made, e.g., if the approach was only tested on a few datasets or with a few runs. In general, empirical results often depend on implicit assumptions, which should be articulated.
        \item The authors should reflect on the factors that influence the performance of the approach. For example, a facial recognition algorithm may perform poorly when image resolution is low or images are taken in low lighting. Or a speech-to-text system might not be used reliably to provide closed captions for online lectures because it fails to handle technical jargon.
        \item The authors should discuss the computational efficiency of the proposed algorithms and how they scale with dataset size.
        \item If applicable, the authors should discuss possible limitations of their approach to address problems of privacy and fairness.
        \item While the authors might fear that complete honesty about limitations might be used by reviewers as grounds for rejection, a worse outcome might be that reviewers discover limitations that aren't acknowledged in the paper. The authors should use their best judgment and recognize that individual actions in favor of transparency play an important role in developing norms that preserve the integrity of the community. Reviewers will be specifically instructed to not penalize honesty concerning limitations.
    \end{itemize}

\item {\bf Theory assumptions and proofs}
    \item[] Question: For each theoretical result, does the paper provide the full set of assumptions and a complete (and correct) proof?
    \item[] Answer: \answerNA{} 
    \item[] Justification: The paper does not include theoretical results.
    \item[] Guidelines:
    \begin{itemize}
        \item The answer NA means that the paper does not include theoretical results. 
        \item All the theorems, formulas, and proofs in the paper should be numbered and cross-referenced.
        \item All assumptions should be clearly stated or referenced in the statement of any theorems.
        \item The proofs can either appear in the main paper or the supplemental material, but if they appear in the supplemental material, the authors are encouraged to provide a short proof sketch to provide intuition. 
        \item Inversely, any informal proof provided in the core of the paper should be complemented by formal proofs provided in appendix or supplemental material.
        \item Theorems and Lemmas that the proof relies upon should be properly referenced. 
    \end{itemize}

    \item {\bf Experimental result reproducibility}
    \item[] Question: Does the paper fully disclose all the information needed to reproduce the main experimental results of the paper to the extent that it affects the main claims and/or conclusions of the paper (regardless of whether the code and data are provided or not)?
    \item[] Answer: \answerYes{}{} 
    \item[] Justification: We provide information on the full model architecture and training objective in \Cref{sec:method}. 
    \Cref{appx:implementation_details} includes further training details. \Cref{sec:experiments} and \Cref{appx:experimental_details} describe recipe for experiments reproduction.
    \item[] Guidelines:
    \begin{itemize}
        \item The answer NA means that the paper does not include experiments.
        \item If the paper includes experiments, a No answer to this question will not be perceived well by the reviewers: Making the paper reproducible is important, regardless of whether the code and data are provided or not.
        \item If the contribution is a dataset and/or model, the authors should describe the steps taken to make their results reproducible or verifiable. 
        \item Depending on the contribution, reproducibility can be accomplished in various ways. For example, if the contribution is a novel architecture, describing the architecture fully might suffice, or if the contribution is a specific model and empirical evaluation, it may be necessary to either make it possible for others to replicate the model with the same dataset, or provide access to the model. In general. releasing code and data is often one good way to accomplish this, but reproducibility can also be provided via detailed instructions for how to replicate the results, access to a hosted model (e.g., in the case of a large language model), releasing of a model checkpoint, or other means that are appropriate to the research performed.
        \item While NeurIPS does not require releasing code, the conference does require all submissions to provide some reasonable avenue for reproducibility, which may depend on the nature of the contribution. For example
        \begin{enumerate}
            \item If the contribution is primarily a new algorithm, the paper should make it clear how to reproduce that algorithm.
            \item If the contribution is primarily a new model architecture, the paper should describe the architecture clearly and fully.
            \item If the contribution is a new model (e.g., a large language model), then there should either be a way to access this model for reproducing the results or a way to reproduce the model (e.g., with an open-source dataset or instructions for how to construct the dataset).
            \item We recognize that reproducibility may be tricky in some cases, in which case authors are welcome to describe the particular way they provide for reproducibility. In the case of closed-source models, it may be that access to the model is limited in some way (e.g., to registered users), but it should be possible for other researchers to have some path to reproducing or verifying the results.
        \end{enumerate}
    \end{itemize}

\item {\bf Open access to data and code}
    \item[] Question: Does the paper provide open access to the data and code, with sufficient instructions to faithfully reproduce the main experimental results, as described in supplemental material?
    \item[] Answer: \answerYes{} 
    \item[] Justification: \Cref{appx:implementation_details} provides links to access source code, trained model, and data for result reproduction. In addition, the code, trained model, and data will be released publicly.
    \item[] Guidelines:
    \begin{itemize}
        \item The answer NA means that paper does not include experiments requiring code.
        \item Please see the NeurIPS code and data submission guidelines (\url{https://nips.cc/public/guides/CodeSubmissionPolicy}) for more details.
        \item While we encourage the release of code and data, we understand that this might not be possible, so “No” is an acceptable answer. Papers cannot be rejected simply for not including code, unless this is central to the contribution (e.g., for a new open-source benchmark).
        \item The instructions should contain the exact command and environment needed to run to reproduce the results. See the NeurIPS code and data submission guidelines (\url{https://nips.cc/public/guides/CodeSubmissionPolicy}) for more details.
        \item The authors should provide instructions on data access and preparation, including how to access the raw data, preprocessed data, intermediate data, and generated data, etc.
        \item The authors should provide scripts to reproduce all experimental results for the new proposed method and baselines. If only a subset of experiments are reproducible, they should state which ones are omitted from the script and why.
        \item At submission time, to preserve anonymity, the authors should release anonymized versions (if applicable).
        \item Providing as much information as possible in supplemental material (appended to the paper) is recommended, but including URLs to data and code is permitted.
    \end{itemize}

\item {\bf Experimental setting/details}
    \item[] Question: Does the paper specify all the training and test details (e.g., data splits, hyperparameters, how they were chosen, type of optimizer, etc.) necessary to understand the results?
    \item[] Answer: \answerYes{} 
    \item[] Justification: All experimental settings are clearly specified. Details of data construction and splits are in \Cref{appx:dataset}, hyperparameters, how they were chosen, and other training details are in \Cref{appx:implementation_details}, and experimental details including evaluation configurations are in \Cref{appx:experimental_details}.
    \item[] Guidelines:
    \begin{itemize}
        \item The answer NA means that the paper does not include experiments.
        \item The experimental setting should be presented in the core of the paper to a level of detail that is necessary to appreciate the results and make sense of them.
        \item The full details can be provided either with the code, in appendix, or as supplemental material.
    \end{itemize}

\item {\bf Experiment statistical significance}
    \item[] Question: Does the paper report error bars suitably and correctly defined or other appropriate information about the statistical significance of the experiments?
    \item[] Answer: \answerNo{} 
    \item[] Justification: For generalist models, reporting results with such a statistical significance demands substantial resources. 
    We note that the prior work we discussed in \Cref{sec:related_works} similarly does not report statistical significance.
    \item[] Guidelines:
    \begin{itemize}
        \item The answer NA means that the paper does not include experiments.
        \item The authors should answer "Yes" if the results are accompanied by error bars, confidence intervals, or statistical significance tests, at least for the experiments that support the main claims of the paper.
        \item The factors of variability that the error bars are capturing should be clearly stated (for example, train/test split, initialization, random drawing of some parameter, or overall run with given experimental conditions).
        \item The method for calculating the error bars should be explained (closed form formula, call to a library function, bootstrap, etc.)
        \item The assumptions made should be given (e.g., Normally distributed errors).
        \item It should be clear whether the error bar is the standard deviation or the standard error of the mean.
        \item It is OK to report 1-sigma error bars, but one should state it. The authors should preferably report a 2-sigma error bar than state that they have a 96\% CI, if the hypothesis of Normality of errors is not verified.
        \item For asymmetric distributions, the authors should be careful not to show in tables or figures symmetric error bars that would yield results that are out of range (e.g. negative error rates).
        \item If error bars are reported in tables or plots, The authors should explain in the text how they were calculated and reference the corresponding figures or tables in the text.
    \end{itemize}

\item {\bf Experiments compute resources}
    \item[] Question: For each experiment, does the paper provide sufficient information on the computer resources (type of compute workers, memory, time of execution) needed to reproduce the experiments?
    \item[] Answer: \answerYes{} 
    \item[] Justification: \Cref{appx:resources} provides sufficient resource information to reproduce the experiments, including workers, memory, and time of execution.
    \item[] Guidelines:
    \begin{itemize}
        \item The answer NA means that the paper does not include experiments.
        \item The paper should indicate the type of compute workers CPU or GPU, internal cluster, or cloud provider, including relevant memory and storage.
        \item The paper should provide the amount of compute required for each of the individual experimental runs as well as estimate the total compute. 
        \item The paper should disclose whether the full research project required more compute than the experiments reported in the paper (e.g., preliminary or failed experiments that didn't make it into the paper). 
    \end{itemize}
    
\item {\bf Code of ethics}
    \item[] Question: Does the research conducted in the paper conform, in every respect, with the NeurIPS Code of Ethics \url{https://neurips.cc/public/EthicsGuidelines}?
    \item[] Answer: \answerYes{} 
    \item[] Justification: The research presented in this paper fully conforms to the NeurIPS Code of Ethics.
    \item[] Guidelines:
    \begin{itemize}
        \item The answer NA means that the authors have not reviewed the NeurIPS Code of Ethics.
        \item If the authors answer No, they should explain the special circumstances that require a deviation from the Code of Ethics.
        \item The authors should make sure to preserve anonymity (e.g., if there is a special consideration due to laws or regulations in their jurisdiction).
    \end{itemize}

\item {\bf Broader impacts}
    \item[] Question: Does the paper discuss both potential positive societal impacts and negative societal impacts of the work performed?
    \item[] Answer: \answerYes{} 
    \item[] Justification: We discuss broader impacts in~\Cref{appx:broader_impacts}.
    \item[] Guidelines:
    \begin{itemize}
        \item The answer NA means that there is no societal impact of the work performed.
        \item If the authors answer NA or No, they should explain why their work has no societal impact or why the paper does not address societal impact.
        \item Examples of negative societal impacts include potential malicious or unintended uses (e.g., disinformation, generating fake profiles, surveillance), fairness considerations (e.g., deployment of technologies that could make decisions that unfairly impact specific groups), privacy considerations, and security considerations.
        \item The conference expects that many papers will be foundational research and not tied to particular applications, let alone deployments. However, if there is a direct path to any negative applications, the authors should point it out. For example, it is legitimate to point out that an improvement in the quality of generative models could be used to generate deepfakes for disinformation. On the other hand, it is not needed to point out that a generic algorithm for optimizing neural networks could enable people to train models that generate Deepfakes faster.
        \item The authors should consider possible harms that could arise when the technology is being used as intended and functioning correctly, harms that could arise when the technology is being used as intended but gives incorrect results, and harms following from (intentional or unintentional) misuse of the technology.
        \item If there are negative societal impacts, the authors could also discuss possible mitigation strategies (e.g., gated release of models, providing defenses in addition to attacks, mechanisms for monitoring misuse, mechanisms to monitor how a system learns from feedback over time, improving the efficiency and accessibility of ML).
    \end{itemize}
    
\item {\bf Safeguards}
    \item[] Question: Does the paper describe safeguards that have been put in place for responsible release of data or models that have a high risk for misuse (e.g., pretrained language models, image generators, or scraped datasets)?
    \item[] Answer: \answerNA{} 
    \item[] Justification: Our model, Mol-LLM, does not have a high risk for misuse.
    \item[] Guidelines:
    \begin{itemize}
        \item The answer NA means that the paper poses no such risks.
        \item Released models that have a high risk for misuse or dual-use should be released with necessary safeguards to allow for controlled use of the model, for example by requiring that users adhere to usage guidelines or restrictions to access the model or implementing safety filters. 
        \item Datasets that have been scraped from the Internet could pose safety risks. The authors should describe how they avoided releasing unsafe images.
        \item We recognize that providing effective safeguards is challenging, and many papers do not require this, but we encourage authors to take this into account and make a best faith effort.
    \end{itemize}

\item {\bf Licenses for existing assets}
    \item[] Question: Are the creators or original owners of assets (e.g., code, data, models), used in the paper, properly credited and are the license and terms of use explicitly mentioned and properly respected?
    \item[] Answer: \answerYes{} 
    \item[] Justification: The backbone models of choice, including LLM and GNN, are credited in \Cref{subsec:architecture}, and the datasets used are properly credited through \Cref{subsec:experimental_setup} and \Cref{appx:dataset}.
    \item[] Guidelines:
    \begin{itemize}
        \item The answer NA means that the paper does not use existing assets.
        \item The authors should cite the original paper that produced the code package or dataset.
        \item The authors should state which version of the asset is used and, if possible, include a URL.
        \item The name of the license (e.g., CC-BY 4.0) should be included for each asset.
        \item For scraped data from a particular source (e.g., website), the copyright and terms of service of that source should be provided.
        \item If assets are released, the license, copyright information, and terms of use in the package should be provided. For popular datasets, \url{paperswithcode.com/datasets} has curated licenses for some datasets. Their licensing guide can help determine the license of a dataset.
        \item For existing datasets that are re-packaged, both the original license and the license of the derived asset (if it has changed) should be provided.
        \item If this information is not available online, the authors are encouraged to reach out to the asset's creators.
    \end{itemize}

\item {\bf New assets}
    \item[] Question: Are new assets introduced in the paper well documented and is the documentation provided alongside the assets?
    \item[] Answer: \answerYes{} 
    \item[] Justification: We release our model, code, and data along with comprehensive documentation, which is available in the code repository as detailed in \Cref{appx:implementation_details}.
    \item[] Guidelines:
    \begin{itemize}
        \item The answer NA means that the paper does not release new assets.
        \item Researchers should communicate the details of the dataset/code/model as part of their submissions via structured templates. This includes details about training, license, limitations, etc. 
        \item The paper should discuss whether and how consent was obtained from people whose asset is used.
        \item At submission time, remember to anonymize your assets (if applicable). You can either create an anonymized URL or include an anonymized zip file.
    \end{itemize}

\item {\bf Crowdsourcing and research with human subjects}
    \item[] Question: For crowdsourcing experiments and research with human subjects, does the paper include the full text of instructions given to participants and screenshots, if applicable, as well as details about compensation (if any)? 
    \item[] Answer: \answerNA{} 
    \item[] Justification: This study does not require human subjects.
    \item[] Guidelines:
    \begin{itemize}
        \item The answer NA means that the paper does not involve crowdsourcing nor research with human subjects.
        \item Including this information in the supplemental material is fine, but if the main contribution of the paper involves human subjects, then as much detail as possible should be included in the main paper. 
        \item According to the NeurIPS Code of Ethics, workers involved in data collection, curation, or other labor should be paid at least the minimum wage in the country of the data collector. 
    \end{itemize}

\item {\bf Institutional review board (IRB) approvals or equivalent for research with human subjects}
    \item[] Question: Does the paper describe potential risks incurred by study participants, whether such risks were disclosed to the subjects, and whether Institutional Review Board (IRB) approvals (or an equivalent approval/review based on the requirements of your country or institution) were obtained?
    \item[] Answer: \answerNA{} 
    \item[] Justification: This study does not require human subjects.
    \item[] Guidelines:
    \begin{itemize}
        \item The answer NA means that the paper does not involve crowdsourcing nor research with human subjects.
        \item Depending on the country in which research is conducted, IRB approval (or equivalent) may be required for any human subjects research. If you obtained IRB approval, you should clearly state this in the paper. 
        \item We recognize that the procedures for this may vary significantly between institutions and locations, and we expect authors to adhere to the NeurIPS Code of Ethics and the guidelines for their institution. 
        \item For initial submissions, do not include any information that would break anonymity (if applicable), such as the institution conducting the review.
    \end{itemize}

\item {\bf Declaration of LLM usage}
    \item[] Question: Does the paper describe the usage of LLMs if it is an important, original, or non-standard component of the core methods in this research? Note that if the LLM is used only for writing, editing, or formatting purposes and does not impact the core methodology, scientific rigorousness, or originality of the research, declaration is not required.
    \item[] Answer: \answerYes{} 
    \item[] Justification: Large language models (LLMs) are an integral component of the proposed method, as described in \Cref{sec:method}. \Cref{sec:method} and \Cref{appx:implementation_details} detail the implementation aspects of the method involving LLMs, while \Cref{appx:baseline_models} provides information on the usage of other baseline LLMs.
    \item[] Guidelines:
    \begin{itemize}
        \item The answer NA means that the core method development in this research does not involve LLMs as any important, original, or non-standard components.
        \item Please refer to our LLM policy (\url{https://neurips.cc/Conferences/2025/LLM}) for what should or should not be described.
    \end{itemize}

\end{enumerate}

\end{document}